\documentclass[10pt,twocolumn,letterpaper]{article}

\usepackage[pagenumbers]{cvpr}  % To force page numbers, e.g. for an arXiv version

\usepackage{graphicx}
\usepackage{capt-of}
\usepackage{xcolor}
\usepackage{amsmath,amssymb,amsfonts,dsfont,pifont,bm,bbm,mathrsfs,mathtools,nicefrac}
\usepackage{algorithm,algpseudocode,listings}
\usepackage{booktabs,multirow,adjustbox,diagbox,threeparttable}
\definecolor{citeblue}{RGB}{48,111,186}
\usepackage[pagebackref,breaklinks,colorlinks,citecolor=citeblue,bookmarks=false]{hyperref}
\usepackage[capitalize]{cleveref}  % Should be loaded after 'hyperref', and works perfectly with 'subfigure'.
\crefname{section}{Sec.}{Secs.}
\Crefname{section}{Section}{Sections}
\crefname{table}{Tab.}{Tabs.}
\Crefname{table}{Table}{Tables}
\crefname{figure}{Fig.}{Figs.}
\Crefname{figure}{Figure}{Figures}
\crefname{equation}{Eq.}{Eqs.}
\Crefname{equation}{Equation}{Equations}
\crefname{conjecture}{Conject.}{Conject.}
\newcommand{\tocite}[1]{\textcolor{red}{[TO CITE]}}

\newcommand{\ieno}{\textit{i}.\textit{e}.}

\def\cvprPaperID{1454}
\def\confName{CVPR}
\def\confYear{2023}

\usepackage{amsmath,amsfonts,bm}

\def\eqref#1{equation~\ref{#1}}
\def\1{\bm{1}}

\def\vzero{{\bm{0}}}
\def\vone{{\bm{1}}}
\def\vmu{{\bm{\mu}}}
\def\vtheta{{\bm{\theta}}}
\def\valpha{{\bm{\alpha}}}

\def\vb{{\bm{b}}}

\def\ve{{\bm{e}}}

\def\vq{{\bm{q}}}

\def\vv{{\bm{v}}}

\def\vx{{\bm{x}}}
\def\vy{{\bm{y}}}

\def\mA{{\bm{A}}}

\def\mQ{{\bm{Q}}}

\def\mU{{\bm{U}}}

\def\mX{{\bm{X}}}

\def\mSigma{{\bm{\Sigma}}}

\DeclareMathAlphabet{\mathsfit}{\encodingdefault}{\sfdefault}{m}{sl}
\SetMathAlphabet{\mathsfit}{bold}{\encodingdefault}{\sfdefault}{bx}{n}

\def\gR{{\mathcal{R}}}

\def\sR{{\mathbb{R}}}

\newcommand{\pdata}{p_{\rm{data}}}
\newcommand{\E}{\mathbb{E}}

\newcommand{\R}{\mathbb{R}}

\newcommand{\Cov}{\mathrm{Cov}}
\begin{document}

% \title{Neural Dependencies of Learning Massive Categories}
\title{Neural Dependencies Emerging from Learning Massive Categories}

\author{Ruili Feng$^1$, Kecheng Zheng$^{2,1}$, Kai Zhu$^1$, Yujun Shen$^2$, Jian Zhao $^1$, Yukun Huang$^1$, Deli Zhao$^{3}$, \\Jingren Zhou$^3$,
Michael Jordan$^4$, Zheng-Jun Zha$^1$\\
$^1$University of Science and Technology of China, Hefei, China\\
$^2$Ant Research, $^3$Alibaba Group, Hangzhou, China\\
$^4$University of California, Berkeley\\
  \texttt{ruilifengustc@gmail.com,}\texttt{\{zkcys001,kaizhu\}@mail.ustc.edu.cn,}\\
  \texttt{shenyujun0302@gmail.com,}
  \texttt{\{zj140,kevinh\}@mail.ustc.edu.cn,}
  \\\texttt{zhaodeli@gmail.com,}\texttt{jingren.zhou@alibaba-inc.com,}\\
  \texttt{jordan@cs.berkeley.edu,}\texttt{zhazj@ustc.edu.cn.} \\
  }

\twocolumn[{
\renewcommand\twocolumn[1][]{#1}
\maketitle
\begin{center}
    \includegraphics[width=0.96\linewidth]{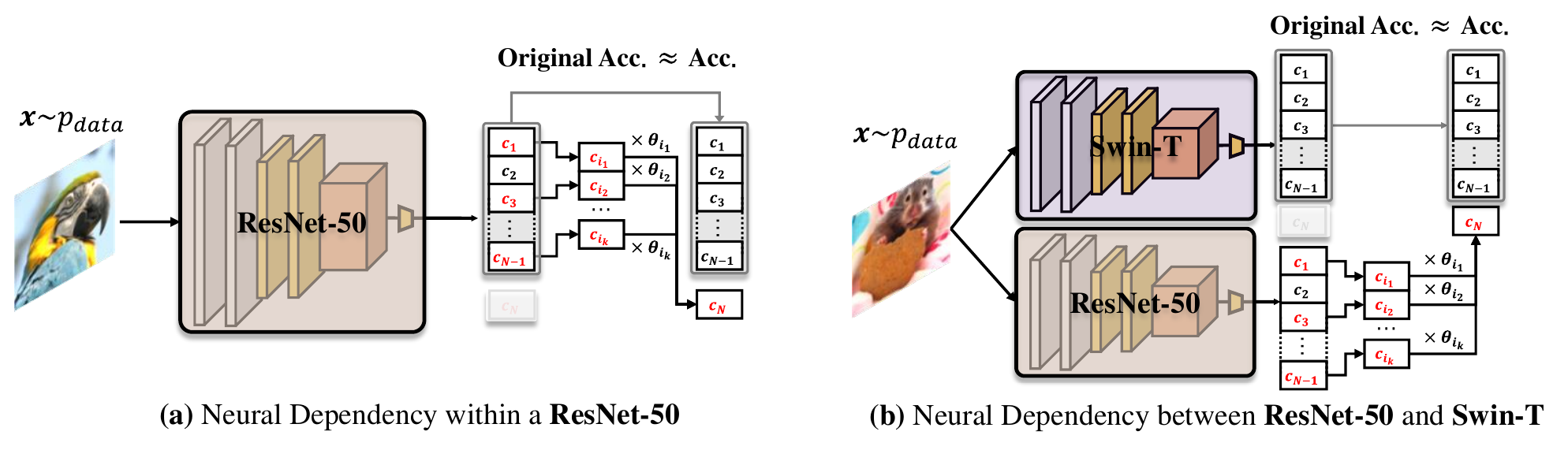}
    \vspace{-5pt}
    \captionsetup{type=figure}
    \caption{
        \textbf{Illustration of neural dependencies} that emerge (a) within a single network and (b) between two independently learned networks.
        Taking the intra-network dependency as an instance, the logits predicted for the category ``macaw'' can be \textit{safely replaced} by a linear combination of the logits predicted for a few other categories, barely scarifying the accuracy.
        %
        % \todo{Add (a)(b), and narrow the figure vertically.}
    }
    \label{fig:teaser}
    \vspace{10pt}
\end{center}
}]

% \maketitle

\begin{abstract}

This work presents two astonishing findings on neural networks learned for large-scale image classification.
1) Given a well-trained model, the logits predicted for some category can be directly obtained by linearly combining the predictions of a few other categories, which we call \textbf{neural dependency}.
2) Neural dependencies exist not only within a single model, but even between two independently learned models, regardless of their architectures.
Towards a theoretical analysis of such phenomena, we demonstrate that identifying neural dependencies is equivalent to solving the Covariance Lasso (CovLasso) regression problem proposed in this paper.
Through investigating the properties of the problem solution, we confirm that neural dependency is guaranteed by a redundant logit covariance matrix, which condition is easily met given massive categories, and that neural dependency is highly sparse, implying that one category correlates to only a few others.
We further empirically show the potential of neural dependencies in understanding internal data correlations, generalizing models to unseen categories, and improving model robustness with a dependency-derived regularizer.
Code for this work will be made publicly available.

\end{abstract}

\vspace{0pt}
\section{Introduction}\label{sec:intro}

Despite the tremendous success of deep neural networks in recognizing massive categories of objects~\cite{imagenet,krizhevsky2017imagenet,he2016deep,dosovitskiy2020image,liu2021swin,krizhevsky2009learning,massive1,massive2,massive3,massive4}, how they manage to organize and relate different categories remains less explored.
A proper analysis of such a problem is beneficial to understanding the network behavior, which further facilitates a better utilization of this powerful tool.

In this work, we reveal that a deep model tends to make its own way of data exploration, which sometimes contrasts sharply with human consciousness.
We reveal some underlying connections between the predictions from a well-learned image classification model, which appears as one category highly depending on a few others.
In the example given in \cref{fig:teaser}a, we can directly replace the logits predicted for ``macaw'' with a linear combination of the logits for ``ostrich'', ``bittern'', \textit{etc.} (without tuning the network parameters) and achieve similar performance.
We call this phenomenon as \textit{neural dependency}, which automatically emerges from learning massive categories.
A more surprising finding is that neural dependencies exist not only within a single model, but also between two independently learned models, as shown in \cref{fig:teaser}b.
It is noteworthy that these two models can even have different architectures (\textit{e.g.}, one with convolutional neural network~\cite{he2016deep} and the other with transformer~\cite{dosovitskiy2020image,liu2021swin}) and different training strategies.

Towards figuring out what brings neural dependencies and whether they happen accidentally, we make a theoretical investigation and confirm that identifying neural dependencies is equivalent to solving a carefully designed convex optimization---the Covariance Lasso (CovLasso) regression problem proposed in this paper.
Such a problem owns a smooth solution path when varying its hyper-parameters~\cite{tibshirani2011solution}, which has two appealing properties.
First, the solution is guaranteed by a redundant covariance matrix of the category-wise logits.
%
%In other words, \check{the redundancy of network predictions} ensures the existence of neural dependencies, which
This condition is easily met when the model is trained on a sufficiently large number of categories~\cite{feng2022rank}.
Second, the solution admits elegant sparsity.
It implies that a category involved in neural dependencies only relates to several instead of numerous other categories.

We further study the potential utilities of neural dependencies, as a support to our theoretical contributions.
One straightforward application is to help interpret the internal data correlations, such as what categories are more likely to link to each other (\cref{subsec:data-analysis}).
Another application is to investigate how we can generalize a well-learned model to unseen categories with the help of neural dependencies (\cref{subsec:generalizability}).
We also propose a regularizer to test whether discouraging the neural dependencies could assist the model in learning a more robust representation (\cref{subsec:robustness}).
We believe the findings in this paper would deepen our understandings on the working mechanism of deep neural networks, and also shed light on some common rules in knowledge learning with visual intelligence systems.

\section{Neural Dependencies}\label{sec:method}
\newcommand{\hCov}{\hat{\Cov}}
\newcommand{\hvtheta}{\hat{\vtheta}}
\newcommand{\hvb}{\hat{\vb}}
\newcommand{\lammax}{\lambda_{\rm{max}}}
\newtheorem{remark}{Remark}
\newtheorem{definition}{Definition}
\newtheorem{theorem}{Theorem}

We consider the $n$-category classification neural network $f:\mathbb{R}^m\rightarrow\mathbb{R}^n$, which takes an input image $\vx\in\mathbb{R}^m$ and outputs the logits vector of $\vx$ being any of the $n$-categories of the task. We assume the network is well-trained and produce meaningful outputs for each category. Naively, each element of the logits vector reports the confidence of the network predicting $\vx$ belonging to the corresponding category. We are curious about whether those confidences can be used to predict each other. Before we start, we formally introduce the key concept of neural dependency in this work.
\begin{definition}\label{def:neural_correlates}
We say the target category $c_i$ and categories $\{c_{i_j}\}_{j=1}^k$ have neural dependency, if and only if for almost every $\vx\sim\pdata$, there are $0<\epsilon,\delta\ll1$ and a few constant non-zero coefficients $\{\vtheta_{i_j}\}_{j=1}^k$, $i\neq i_j\in[n],k\ll n$, such that
\begin{equation}
    \Pr(\vert f(\vx)_i-\sum_{j=1}^k\vtheta_{i_j}f(\vx)_{i_j}\vert<\epsilon)>1-\delta.
\end{equation}
\end{definition}
\begin{remark}
We do not normalize nor centralize the logits output $f(\vx)$ so that no information is added or removed for logits of each category. Different from usual linear dependency system (where $\vy=\mA \vx+\vb$), we omit bias in the neural dependency, \ieno, we require $b=0$ if $f(\vx)_i\approx \sum_{j=1}^k\vtheta_{i_j}f(\vx)_{i_j}+b$. Thus the existence of neural dependencies suggests that the network believes category $c_i$ is nearly purely decided by categories $c_{i_1},\cdots,c_{i_k}$ without its own unique information.
\end{remark}
\paragraph{What Means When We Have Neural Dependencies?} The neural dependency means that a linear combination of a few categories is in fact another category. It is natural to believe that those categories should admit certain intrinsic correlations. However, for an idea classifier, each category should hold a unique piece of information thus they should not be purely decided by other categories. What's more, we will find that some neural dependencies are also not that understandable for humans. Overall, the neural dependencies reveal a rather strong intrinsic connection of hidden units of neural networks, and are potentially interesting for understanding the generality and robustness of networks.

\newcommand{\tCov}{\tilde{\Cov}}
\newcommand{\tf}{\tilde{f}}
\paragraph{Between Network Dependencies.} 
We can also solve and analyze the between network neural dependencies through the above methodology for two different neural networks $f,g$ trained on the same dataset independently. Here we want to find a few constant non-zero coefficients $\{\vtheta_{i_j}\}_{j=1}^k$ such that $\Pr(\vert g(\vx)_i-\sum_{j=1}^k\vtheta_{i_j}f(\vx)_{i_j}\vert<\epsilon)>1-\delta$. To find those coefficients, we only need to use the $i$-th row of $g(\vx)$ to replace $f(\vx)_i$ in \cref{problem:cLasso}. The concepts of within and between network dependencies are also illustrated in \cref{fig:teaser}.

\paragraph{Notations.} We use bold characters to denote vectors and matrix, under-subscript to denote their rows and upper-subscript to denote their columns. For example, for a matrix $\vmu$, $\vmu_{A}^B$ denote the sub-matrix of $\vmu$ consists of the elements with row indexes in set $A$ and column indexes in set $B$; for a vector $\vtheta$, we use $\vtheta_i$ to denote its $i$-th row which is a scalar. For a function $f:\R^m\rightarrow\R^n$, $f(\vx)_i$ denote the $i$-th row of vector $f(\vx)$, while $f_i(\vx)$ is some other function that connected with sub-script $i$. For an integer $n\in\mathbb{N}$, we use $[n]$ to denote the set $\{1,\cdots,n\}$. We always assume that matrices have full rank unless specifically mentioned; low-rank matrices are represented as full rank matrices with many tiny singular values (or eigenvalues for symmetry low-rank matrices).

\paragraph{Experiments Setup in This Section.} In this section we reveal the neural dependencies empirically among some most popular neural networks, \ieno, ResNet-18, ResNet-50~\cite{he2016deep}, Swin-Transformer~\cite{liu2021swin}, and Vision-Transformer~\cite{dosovitskiy2020image}. As a benchmark for massive category classification, we use ImageNet-1k~\cite{imagenet}, which includes examples ranging from 1,000 diverse categories, as the default dataset. Training details of those networks, and other necessary hyper-parameters to reproduce the results in this paper can be found in the appendix.

\subsection{Identifying Neural Dependencies through Covariance Lasso}
We propose the Covariance Lasso (CovLasso) problem which will help us identify the neural dependencies in the network and play an essential role in this paper:
\begin{equation}\label{problem:cLasso}
    \min_{\vtheta\in\R^n,\vtheta_i=-1}\mathbb{E}_{\vx\sim\pdata}[\Vert\vtheta^T f(\vx)\Vert_2^2]+\lambda\Vert\vtheta\Vert_1.
\end{equation}
Let $\vtheta^*(\lambda)$ be the solution of \cref{problem:cLasso} given hyper-parameter $\lambda>0$, we can have the following observations
\begin{enumerate}
    \item $\vtheta^*(\lambda)$ will be a sparse $n$-dimensional vector, meaning many of its elements will be zero, due to the property of $\ell_1$ penalty~\cite{tibshirani1996regression};
    \item the prediction error $\vert f_i(\vx)-\sum_{k=1}^s\vtheta^*(\lambda)_{i_k}f_{i_k}(\vx)\vert=\Vert\vtheta^{*T}(\lambda)f(\vx)\Vert_2^2$ will be very small for most $\vx\sim\pdata$, due to the property of minimization of expectation.
\end{enumerate}
Combining these two observations, it is easy to find out the solution of \cref{problem:cLasso} naturally induces the linear neural dependencies in \cref{def:neural_correlates}. Rigorously, by Markov inequality, if $\E_{\vx\sim\pdata}[\Vert\vtheta^T f(\vx)\Vert_2^2]\leq\epsilon\delta$, we have
\begin{equation}\label{eq:markov ineq}
\begin{aligned}
         &\Pr(\vert f(\vx)_i-\sum_{j\neq i}\vtheta_j f(\vx)_j\vert <\epsilon) \\
         =&1-\Pr(\vert f(\vx)_i-\sum_{j\neq i}\vtheta_j f(\vx)_j\vert \geq\epsilon)\\
         \geq &1-\frac{\E_{\vx\sim\pdata}[\Vert \vtheta^Tf(\vx)\Vert_2^2]}{\epsilon}\geq 1-\delta,
\end{aligned}
\end{equation}
so we can have the following theorem.
\begin{theorem}\label{th:equi}
The solution to \cref{problem:cLasso} satisfies \cref{def:neural_correlates} for some small $\epsilon$ and $\delta$ and appropriate $\lambda$.
\end{theorem}
The CovLasso problem is a convex problem; we can efficiently solve it by various methods like coordinate descent or subgradient descent~\cite{boyd2004convex}. Finding the neural dependencies for some category $c_i$ is now transferring into solving the CovLasso problem under the constraint $\vtheta_i=-1$. 

\begin{figure*}[t]
    \centering
    \includegraphics[ width=0.96\linewidth]{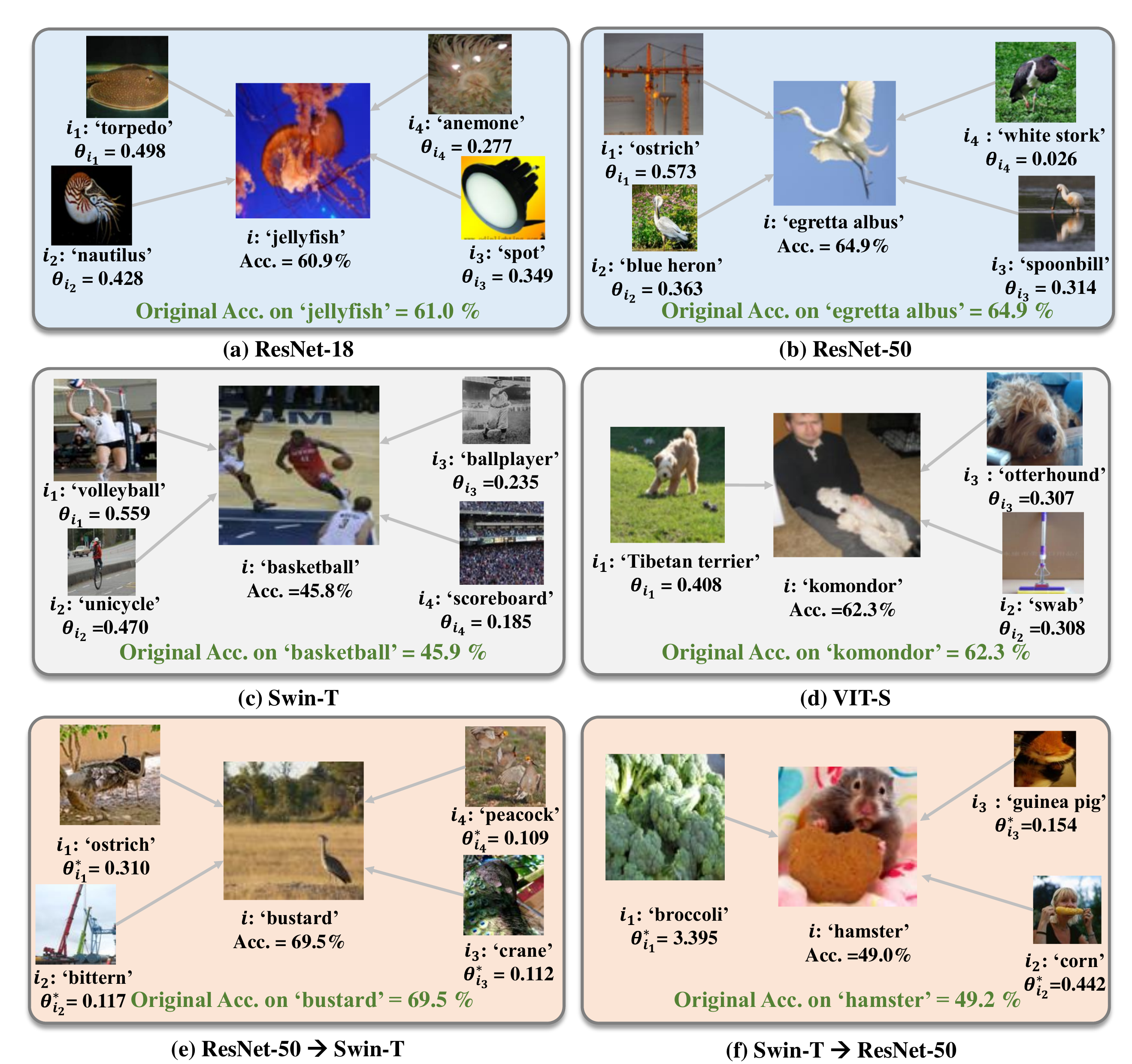}

    \caption{Neural dependencies in popular multi-class classification networks. (a;b;c) Within-network neural dependencies in ResNet18, ResNet50, Swin-Transformer and VIT-Transformer; (e;f) Between-network neural dependencies between ResNet50 and Swin-Transformer. Much more results can be found in appendix.}
    \label{fig:ND_exps}
\end{figure*}

\paragraph{Results.}\cref{fig:ND_exps} reports some results of both within and between network neural dependencies acquired by solving \cref{problem:cLasso}. In the center we report the target category and in the surroundings we enumerate those categories that emerge neural dependencies with it. We show more results in the appendix. For the cases in \cref{fig:ND_exps}, \cref{tab:ND} further reports the absolute and relative errors of predicting the logits of target categories using formula $f(\vx)_i\approx \sum_{k=1}^s\vtheta_{i_k}f(\vx)_{i_k}$, and the corresponding classification accuracy on this category (using the replaced logits $(f(\vx)_1,\cdots,f(\vx)_{i-1},\sum_{j\neq i}\vtheta_j f(\vx)_j,f(\vx)_{i+1},\cdots,f(\vx)_n)^T$ instead of $f(\vx)$), tested both on positive samples only and the full validation set of ImageNet. We can  find that, as claimed by \cref{def:neural_correlates}, a small number of other categories (3 or 4 in the illustrated cases) are enough to accurately predict the network output for the target category. Moreover, the predictions are all linear combinations: for example, \cref{fig:ND_exps} (f) tells that for almost every image $\vx\sim\pdata$, we have
\begin{equation}
\begin{aligned}
    {\rm R50}&(\vx)_{\rm hamster}\approx 3.395\times {\rm S}(\vx)_{\rm broccoli}\\
    &+3.395\times {\rm S}(\vx)_{\rm guinea pig}+3.395\times {\rm S}(\vx)_{\rm corn},
\end{aligned}
\end{equation}
where ${\rm R50}$ denotes the ResNet-50 and ${\rm S}$ denotes the Swin-Transformer. We can achieve comparable classification performance if using the above linear combination to replace the logits output for category `hamster' of ResNet-50. For both single models and two independently trained models with different architectures, we can observe clear neural dependencies. Future work may further investigate connections and differences in neural dependencies from different networks.

\paragraph{Peculiar Neural Dependencies.} As we have mentioned before, the solved neural dependencies are not all that understandable for humans. \cref{fig:ND_exps} actually picks up a few peculiar neural dependencies for both within and between network dependencies. For example, the dependencies between `jellyfish' and `spot' in \cref{fig:ND_exps}a, `egretta albus' and `ostrich' in \cref{fig:ND_exps}b, `basketball' and `unicycle' in \cref{fig:ND_exps}c, `komondor' and `swab' in \cref{fig:ND_exps}d,  `bustard' and `bittern' in \cref{fig:ND_exps}e, and `hamster' and `brocoli' in \cref{fig:ND_exps}f. This reveals the unique way of understanding image data of neural networks compared with human intelligence that has been unclear in the past~\cite{selvaraju2017grad,zhou2016learning}. Further investigating those cases can be of general interests to future works in AI interpretability and learning theory, and potentially provide a new way to dig intrinsic information in image data.

\begin{table*}[]
    \centering
    \caption{Prediction error and classification accuracy of neural dependencies in cases in \cref{fig:ND_exps}. Both the error of logits prediction and the loss in classification accuracy are tiny. Much more results can be found in appendix.}\label{tab:ND}
    \resizebox{\textwidth}{!}
    {
    \renewcommand{\arraystretch}{1.25}
    \begin{tabular}{c||c|c|c|c|c|c}
    \toprule
    Metrics  & ResNet-18   & ResNet-50   & Swin-T      & VIT-S       & R-50 $\to$ Swin-T  & Swin-T $\to$ R-50 \\ \hline
    Abs Err & 0.187       & 0.068       & 0.104     & 0.297      & 0.207                  & 0.211                \\ \hline
    Rel Err (\%) & 2.568       & 1.063       & 0.926      & 4.276      & 1.776                  & 3.939                 \\ \hline
    Acc (Ori. Acc)     & 60.9 (61.0) & 64.9 (64.9) & 40.1 (40.1) & 45.9 (45.9) & 69.5 (69.5)             & 49.0 (49.2)            \\ \hline
    Pos Acc (Ori. Pos Acc) & 72.0 (84.0) & 92.0 (92.0) & 94.0 (92.0) & 96.0 (100.0) & 94.0 (96.0)             & 94.0 (100.0)           \\ \bottomrule
    \end{tabular}
    }
\end{table*}

\subsection{What Brings Dependencies}
After identifying the neural dependencies in deep networks, we are curious about why this intriguing phenomenon can broadly exist in different architectures. So we need a further understanding of the sources of it, which can be discovered through a careful analysis on \cref{problem:cLasso}. This section will reveal how a redundant covariance matrix for the terminal representations induces the neural dependencies.

Observe that $\E_{\vx\sim\pdata}[\Vert\vtheta^T f(\vx)\Vert_2^2]=\vtheta^T\Cov\vtheta$, where $\Cov=\E_{\vx\sim\pdata}[ f(\vx) f(\vx)^T ]$ is the (uncerntralized and unnormalized) covariance matrix of the terminal representations. Let $\bm{err}_i(\vtheta)=\vtheta^T\Cov\vtheta$ be the predicting error of using coefficient $\vtheta$ for category $c_i$, the property of Lasso regression indicates that (see proof in appendix) $\bm{err}_i(\vtheta^*(\lambda))$ is continuous about $\lambda$ and
\begin{equation}\label{eq:err_path}
\begin{aligned}
     \frac{\det[\Cov]}{\det[\Cov_{[n]\setminus i}^{[n]\setminus i}]}& =\bm{err}_i(\vtheta^*(0) \\
    & \leq\bm{err}_i(\vtheta^*(\lambda))\leq\bm{err}_i(\vtheta^*(\lambda')) \\
    & \leq\bm{err}_i(\vtheta^*(\lambda_{\rm{max}}))=\Cov_i^i,
\end{aligned}
\end{equation}
where $\lambda\leq\lambda'$, and $\lambda_{max}=2\Vert\Cov_{[n]\setminus i}^i\Vert_{\infty}$ is the supremum of valid hyper-parameter $\lambda$, \ieno, $\vtheta^*(\lambda)=-\ve_i=(\underbrace{0,\cdots,0}_{i-1},-1,0,\cdots,0),\forall \lambda\geq\lambda_{\rm{max}}$, and $\vtheta^*(\lambda)\neq-\ve_i,\forall0\leq\lambda<\lambda_{\rm{max}}$.

Regardless of the sparsity, to yield neural dependency for the target category $c_i$, we expect a very small $\bm{err}_i(\vtheta^*(\lambda))$. So if the lower bound $\bm{err}_i(\vtheta^*(0))$ is already far larger than $\epsilon\delta$, the predicting error can be too large to yield neural dependencies. Reversely, using the continuity of $\bm{err}_i(\vtheta^*(\lambda))$ about $\lambda$, we can know that if the lower bound $\bm{err}_i(\vtheta^*(0))$ is very small, then there should be a small $\lambda$ such that $\bm{err}_i(\vtheta^*(\lambda))$ is also very small. \cref{problem:cLasso} can then bring neural dependencies to category $c_i$. (This need to exclude a trivial case where the predicting error upper bound $\Cov_i^i=\E_{\vx\sim\pdata}[f(\vx)_i^2]$ is already very small as it does not reveal any meaningful dependencies but that the network may be very unconfident about category $c_i$. While this is rare for well-trained networks, we leave the discussion of this case in appendix.) 

So to have neural dependencies, we require the term $\bm{err}_i(\vtheta^*(0))$ to be as small as possible. For term $\bm{err}_i(\vtheta^*(0))$ we can have the following observations from two different perspectives (see appendix for deduction):
\begin{enumerate}
    \item Information Volume: $\bm{err}_i(\vtheta^*(0))=\frac{\det[\Cov]}{\det[\Cov_{[n]\setminus i}^{[n]\setminus i}]}=\frac{{\rm Vol}(\Cov)}{{\rm Vol}(\Cov_{[n]\setminus i}^{[n]\setminus i})}$ measures the ratio between the $n$-dimensional volumes of the parallelotope $\Cov$ and the $n-1$ dimensional volumes of $\Cov$ removing the $i$-th row and $i$-th column; if assume Gaussian distributions of random variable $f(\vx),\vx\sim\pdata$, they are also the normalizing constants of the probability density of the terminal representations with and without the $i$-th category; this term measures the information loss while removing the $i$-th category and is small if the $i$-th row and $i$-th column of $\Cov$ carry little information and are redundant;
    \item Geometry: $\bm{err}_i(\vtheta^*(0))=\frac{\det[\Cov]}{\det[\Cov_{[n]\setminus i}^{[n]\setminus i}]}=(\sum_{j=1}^n\frac{\valpha_j^2}{\sigma_j^2})^{-1}$ which will be small if some $\valpha_j$ corresponding to tiny $\sigma_j^2$ is large, where $\sigma_1^2\geq\cdots\geq\sigma_n^2$ are the eigenvalues of $\Cov$ and $\vq_1,\cdots,\vq_n$ are the corresponding eigenvectors, $\valpha_j=\langle\ve_i,\vq_j\rangle,j\in[n]$; this further means that the $i$-th coordinate axis is close to the null space (linear subspace spanned by eigenvectors corresponding to tiny eigenvalues) of the covariance matrix $\Cov$, which suggests the $i$-th category is redundant geometrically.
\end{enumerate}

Let $\frac{\det[\Cov]}{\det[\Cov_{[n]\setminus i}^{[n]\setminus i}]}$ be the metric for redundancy of category $c_i$, both perspectives lead to the same conclusion that:
\begin{center}
    \textit{Redundancy of the target category $c_i$ in the terminal representations brings it neural dependencies.}
\end{center}

% \begin{figure}
%     \centering
%     \includegraphics[width=1.0\linewidth]{figs/redundancy_ratio.pdf}
%     \caption{redundancy ratio v.s. prediction error}
%     \label{fig:redundancy}
% \end{figure}
\begin{remark}
Unfortunately, though it can help us understand the intrinsic mechanism that brings neural dependencies, this principle is only intuitive in practice---we can not accurately calculate the value $\frac{\det[\Cov]}{\det[\Cov_{[n]\setminus i}^{[n]\setminus i}]}$ in most cases due to numerical instability. $\det[\Cov_{[n]\setminus i}^{[n]\setminus i}]$ tends to have some tiny singular values (smaller than ${\rm 1e-3}$), making the quotient operation extremely sensitive to minor numerical errors in computation, and thus often induces ${\rm NaN}$ results.
\end{remark}

\subsection{What Brings Sparsity}
The last section omits the discussion of sparsity, which we want to study carefully in this section. We want to find a value that estimates whether two categories have neural dependencies, which we will show later is the (uncerntralized) covariance between the logits for two different categories.

\begin{figure}
    \centering
    \includegraphics[width=1.0\linewidth]{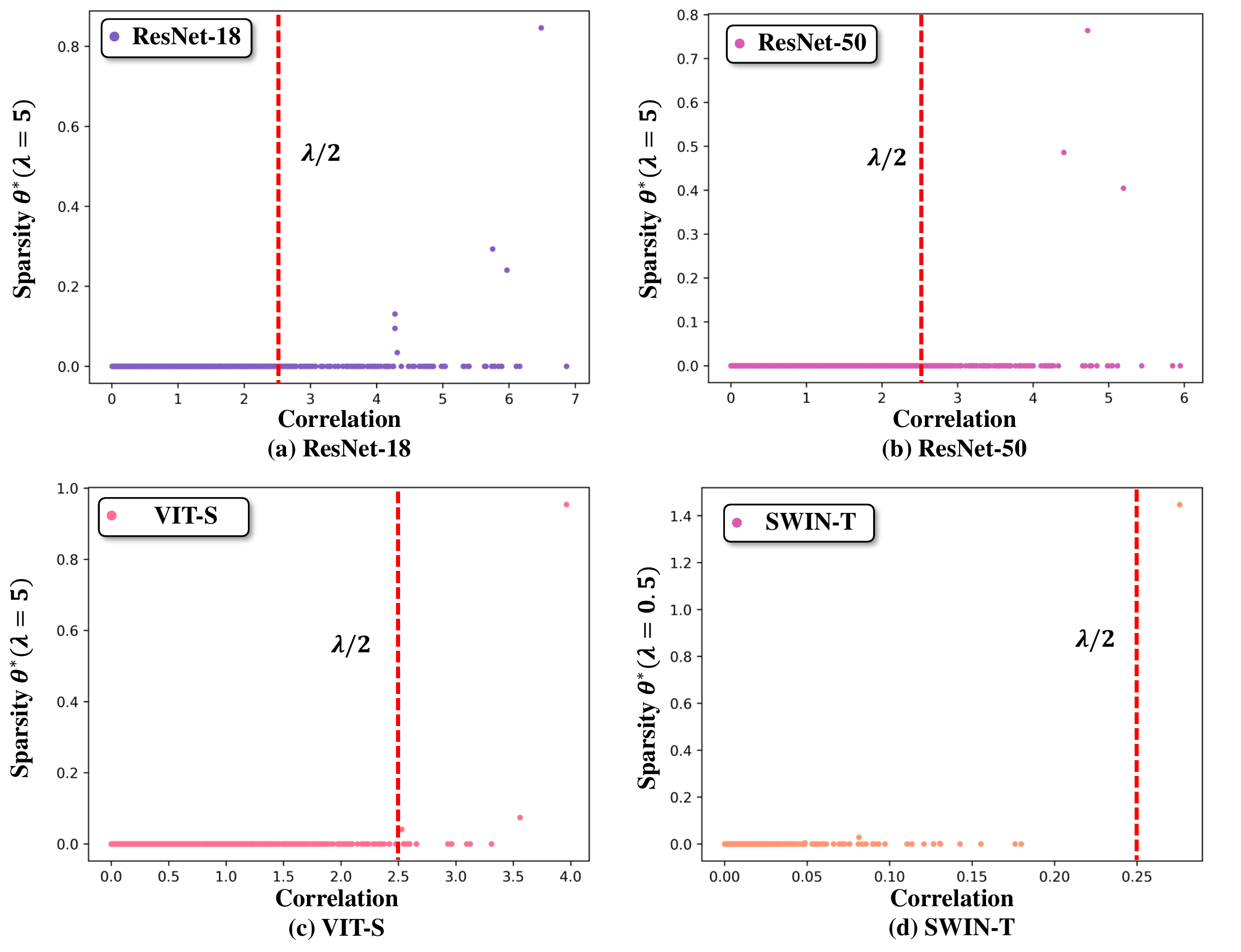}\vspace{-0.3cm}
    \caption{Relation between correlations and dependency coefficients. }\vspace{-0.4cm}
    \label{fig:correlat_sparsity}
\end{figure}

The sparsity property, \ieno, whether category $c_j$ is involved in the neural dependencies with $c_i$, can be identified by the KKT condition of \cref{problem:cLasso}. Let $\hCov=\Cov_{[n]\setminus i}^{[n]\setminus i}$, $\hvtheta=\vtheta_{[n]\setminus i}$, $\hvb=\Cov_{[n]\setminus i}^i$, and $\hat{j}=j+\vone_{(j>i)}$ such that $\hvtheta_{j}=\vtheta_{\hat{j}}$, \cref{problem:cLasso} then transfer into
\begin{equation}\label{problem:classo2}
    \min_{\hvtheta\in\R^{n-1}}\hvtheta^T\hCov\hvtheta-2\hvb^T\hvtheta+\lambda\Vert\hvtheta\Vert_1.
\end{equation}
By KKT conditions, the optimal value is attained only if
\begin{equation}
    \vzero\in\hCov\hvtheta^*(\lambda)-\hvb+\frac{\lambda}{2}\partial\Vert\hvtheta\Vert_1.
\end{equation}
and the sparsity can be estimated by the following proposition (see detailed deduction in appendix)
\begin{gather}
    \vert\hCov_j\hvtheta^*(\lambda)-\hvb_j\vert<\frac{\lambda}{2}\Rightarrow\hvtheta^*(\lambda)_j=0,j\in[n-1].
\end{gather}
This means that we can know whether two categories admit neural dependencies by estimating $\vert\hCov_j\hvtheta^*(\lambda)-\hvb_j\vert$. A surprising fact is that the term $\vert\hCov_j\hvtheta^*(\lambda)-\hvb_j\vert$ can actually be estimated without solving \cref{problem:cLasso}, but using the slope of the solution path of the Lasso problem. By convexity of \cref{problem:cLasso}, the slope of \cref{problem:cLasso} admits the following bound.
\begin{theorem}
Let $\hCov=\mQ\mSigma\mQ^T$ be the eigenvalue decomposition of $\hCov$, and $\mA=\mQ\mSigma^{1/2}\mQ^T$, then we have for $\lambda',\lambda''\in[0,\lambda_{\rm{max}}]$,
\begin{equation}\label{th:slope}
\begin{aligned}
        &\vert\frac{\hCov_j\hvtheta^*(\lambda')-\hvb_j}{\lambda'}-\frac{\hCov_j\hvtheta^*(\lambda'')-\hvb_j}{\lambda''}\vert\\
        \leq &\Vert\mA_j\Vert_2\Vert\mA^{-T}\hvb\Vert_2\vert\frac{1}{\lambda'}-\frac{1}{\lambda''}\vert,j\in[n-1].
\end{aligned}
\end{equation}
\end{theorem}
\begin{remark}
Using this theorem we can also get a finer estimation of the value of $\bm{err}_i(\hvtheta^*(\lambda))$ than \cref{eq:err_path}, see appendix for detail.
\end{remark}
Using triangular inequality and the closed-form solution for $\lambda_{\rm{max}}$ ($\hvtheta^*(\lambda_{\rm_{max}})=\vzero$), we have for $j\in[n-1]$,
\begin{align}\label{eq:pre_screen}
    &\vert\hCov_j\hvtheta^*(\lambda)-\hvb_j\vert
    \leq\lambda\vert\frac{\hCov_j\hvtheta^*(\lambda_{\rm{max}})-\hvb_j}{\lambda_{\rm{max}}}\vert
    \\
    +&\lambda\vert \frac{\hCov_j\hvtheta^*(\lambda)-\hvb_j}{\lambda}-\frac{\hCov_j\hvtheta^*(\lambda_{\rm{max}})-\hvb_j}{\lambda_{\rm{max}}}\vert\\
    \leq &\lambda\vert\frac{\hvb_j}{\lammax}\vert
    +\lambda\Vert\mA_j\Vert_2\Vert\mA^{-T}\hvb\Vert_2\vert\frac{1}{\lambda}-\frac{1}{2\Vert\hvb\Vert_{\infty}}\vert.
\end{align}
Thus if $\lambda\vert\frac{\hvb_j}{\lammax}\vert+\lambda\Vert\mA_j\Vert_2\Vert\mA^{-T}\hvb\Vert_2\vert\frac{1}{\lambda}-\frac{1}{2\Vert\hvb\Vert_{\infty}}\vert<\frac{\lambda}{2}\Leftrightarrow \vert\frac{\hvb_j}{\Vert\hvb\Vert_{\infty}}\vert <1-2\Vert\mA_j\Vert_2\Vert\mA^{-T}\hvb\Vert_2\vert\frac{1}{\lambda}-\frac{1}{2\Vert\hvb\Vert_{\infty}}\vert$, we know that $\hvtheta^*(\lambda)_j=0$ and category $c_{\hat{j}}$ is independent (meaning not involved in the neural dependencies) with $c_i$.
\begin{theorem}\label{th:screening}
When $0<\lambda<\lambda_{\rm{max}}$ and $\hat{j}\neq i$, if
\begin{equation}
\begin{aligned}
    &\frac{\vert\E_{\vx\sim\pdata}[f(\vx)_i f(\vx)_{\hat{j}}]\vert}{\max_{s\neq i}\vert\E_{\vx\sim\pdata}[f(\vx)_i f(\vx)_s]\vert}\\
    <&1-2\Vert\mA_j\Vert_2\Vert\mA^{-T}\hvb\Vert_2\vert\frac{1}{\lambda}-\frac{1}{2\Vert\hvb\Vert_{\infty}}\vert,
\end{aligned}
\end{equation}
then $\vtheta^*(\lambda)_{\hat{j}}=0$ and category $c_{\hat{j}}$ is independent with $c_i$.
\end{theorem}
High dimensional vectors are known to tend to be orthogonal to each other~\cite{buhlmann2011statistics}, thus if we assume $\mA_j$ is nearly orthogonal to $\mA^{-T}\hvb$, then $\Vert\mA_j\Vert_2\Vert\mA^{-T}\hvb\Vert_2\approx\vert\hvb_j\vert$ and we can further simplify the above sparsity criterion as
\newtheorem{conjecture}{Conjecture}
\begin{conjecture}\label{conj:screening}
When $0<\lambda<\lambda_{\rm{max}}$ and $j\neq i$, if
\begin{equation}
\begin{aligned}
        \vert\E_{\vx\sim\pdata}[f(\vx)_i f(\vx)_{j}]\vert<\frac{\lambda}{2}(\text{equivalent to
        }\\
        \frac{\vert\E_{\vx\sim\pdata}[f(\vx)_i f(\vx)_{j}]}{\max_{s\neq i}\vert\E_{\vx\sim\pdata}[f(\vx)_i f(\vx)_s]\vert}\vert<\frac{\lambda}{\lammax}),
\end{aligned}
\end{equation}
then $\vtheta^*(\lambda)_{j}=0$ and category $c_{j}$ is independent with $c_i$.
\end{conjecture}
In practice we find that this conjecture is seldom wrong. Combining with \cref{th:screening}, they together tell us that the covariance of terminal representations has an important role in assigning neural dependencies: more correlated categories tend to have neural dependencies, while weakly correlated categories will not have neural dependencies. They also describe the role of the hyper-parameter $\lambda$ in \cref{problem:cLasso}: it screens out less correlated categories when searching neural dependencies, and larger $\lambda$ corresponds to higher sparsity of dependencies. In conclusion, let $\vert\E_{\vx\sim\pdata}[f(\vx)_i f(\vx)_{j}]\vert$ be the metric for correlations between category $c_i$ and $c_j$, we can say that
\begin{center}
    \textit{Low covariance between categories in the terminal representations brings sparsity of dependencies.}
\end{center}
\paragraph{Numerical Validation.} We validate the above principle, \ieno, \cref{conj:screening} in \cref{fig:correlat_sparsity}. Each subfigure picks up one target category $c_i$ and solves \cref{problem:cLasso} to calculate the corresponding coefficients $\vtheta_j^*,j\neq i$ for all the remaining 999 categories of the ImageNet. $\vtheta^*_j=0$ implies no neural dependency between category $c_i$ and $c_j$, and vice versa. We plot the relation between the covariance of $c_i,c_j$, $\vert\E_{\vx\sim\pdata}[f(\vx)_i f(\vx)_{j}]\vert$, and the corresponding dependency coefficient $\vtheta_j^*$. We can clearly find out that a small correlation corresponds to no neural dependency. Specifically, when the correlation between $c_i,c_j$ is smaller than $\frac{\lambda}{2}$, $c_i$ and $c_j$ admit no neural dependency. In most cases, the bar $\frac{\lambda}{2}$ does exclude a considerable amount of zero dependency categories, which makes it a good indicator for the existence of neural dependency. This validates our principle for the source of sparsity.

\paragraph{Controlling Neural Dependencies.} \cref{conj:screening} also points out that we can disentangle neural dependencies by regularizing the covariance term, as tiny covariance of $\vert\E_{\vx\sim\pdata}[f(\vx)_i f(\vx)_{j}]\vert$ indicates no neural dependency between category $c_i$ and $c_j$. We will discuss this later in \cref{subsec:robustness}.

\subsection{Between Network Neural Dependencies}
 The general math property of the between network neural dependencies shows no essential difference from the within network ones. Let $f,g$ be two different classification neural networks trained on $\pdata$ independently. We want to use the logits of $f$ to predict the logits of the $c_i$ category of $g$.  Let $\tf(\vx)=(f(\vx)_1,\cdots,f(\vx)_{i-1},g(\vx)_i,f(\vx)_{i+1},\cdots,f(\vx)_n)^T$, and $\tCov=\E_{\vx\sim\pdata}[\tf(\vx)\tf(\vx)^T]$, then we know that
\begin{itemize}
    \item if $\frac{\det[\tCov]}{\det[\tCov]_{[n]\setminus i}^{[n]\setminus i}}=\frac{\det[\tCov]}{\det[\Cov_{[n]\setminus i}^{[n]\setminus i}]}$ is small, then category $c_i$ of network $g$ have neural dependencies with some other categories of network $f$;
    \item if $\vert\E_{\vx\sim\pdata}[f(\vx)_jg(\vx)_i]\vert$ ($j\neq i$) is small, then the $c_j$ category of $f$ is independent with the $c_i$ category of $g$.
\end{itemize}

\section{Potentials of Neural Dependencies}\label{sec:appls}
In this section, we discuss some interesting potentials and inspirations of neural dependencies in general scenarios of modern machine learning.

\subsection{Visualization Neural Dependencies}\label{subsec:data-analysis}
\begin{figure}[t]
    \centering
    \includegraphics[ width=1.0\linewidth]{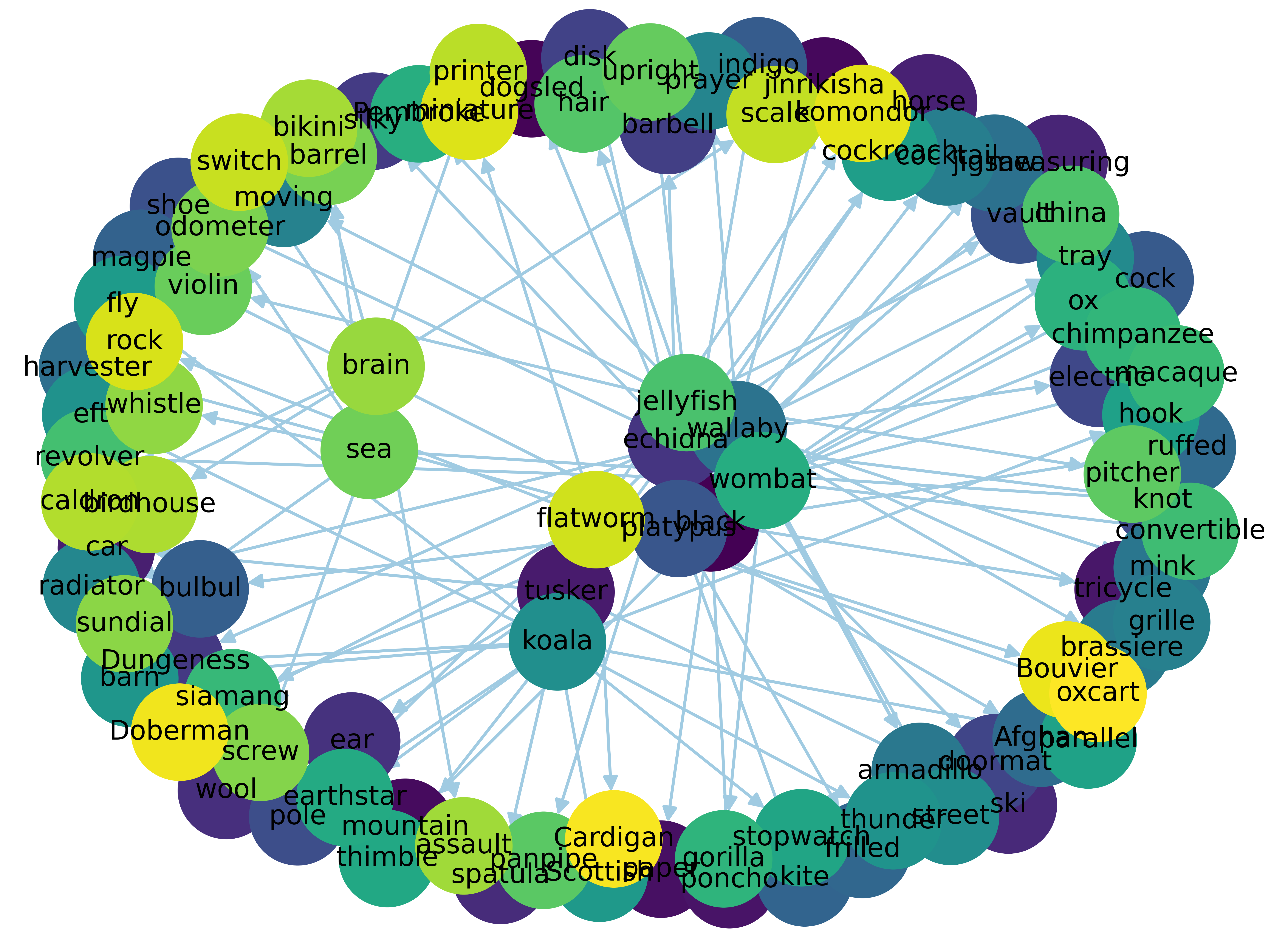}\vspace{-0.2cm}

    \caption{The graph visualization of neural dependencies in a pretrained ResNet-50. Please refer to \cref{subsec:data-analysis} for detail.}
    \label{fig:AD_exps}\vspace{-0.4cm}
\end{figure}

We are curious about the intrinsic data relations revealed by neural dependencies. Specifically, if we have some base classes in the coordinate space, can we plot the position of the target classes that can be linearly decided by (have neural dependencies with) those classes? \cref{fig:AD_exps} gives such an example for ResNet-50 in ImageNet. In the surroundings are 88 base categories and in the center are 10 target categories that can be linearly predicted by them using neural dependencies. The length of the arc between two categories gives their dependency coefficient. This result illustrates the relative relationship of different categories acknowledged by the neural network. It may be of potential interests to multiple domains like data relation mining, visualization, and interpretability of deep networks.

\begin{table*}[ht]
  \renewcommand{\arraystretch}{1.}
  \renewcommand{\tabcolsep}{4.5pt}
    \centering
    \caption{Classification accuracy of baselines and learning new categories through neural dependencies (ours). While much simpler, learning new categories through neural dependencies barely looses accuracy. All figures are the mean of five independent runs.}\vspace{-0.3cm}
    \label{tab:generalization}
    \begin{tabular}{l||ccc|ccc|ccc|ccc}
    \toprule
    % \Xhline{2.\arrayrulewidth}
    \multicolumn{1}{c||}{} & \multicolumn{3}{c|}{\textbf{$900\to100$}}  & \multicolumn{3}{c|}{\textbf{$950\to50$}} & \multicolumn{3}{c|}{\textbf{$999\to1$}} & \multicolumn{3}{c}{\textbf{$999\to1(pos\&neg)$}} \\ \cline{2-4} \cline{5-7} \cline{8-10} \cline{11-13}
    \multicolumn{1}{c||}{\multirow{-2}{*}{\textbf{Backbone}}} & \textit{Baseline} & \textit{Ours} & \textit{Impro} & \textit{Baseline} & \textit{Ours} & \textit{Impro}& \textit{Baseline} & \textit{Ours} & \textit{Impro}& \textit{Baseline} & \textit{Ours} & \textit{Impro}\\ 
    \hline
    ResNet50 &68.47&68.03&-0.44&68.47&68.45&-0.02&68.47&68.46&-0.01&60.70&61.50&+0.80  \\ 
    \hline
    Swin-T   
    &71.49 &71.486 &-0.004 &71.49 &71.578 &+0.088 &71.49 &71.56 &+0.07 & 76.20 & 78.00 & +1.80  \\ 
    \bottomrule
    % \Xhline{2.\arrayrulewidth}
    \end{tabular}
    %\vspace{-0.2cm}
\end{table*}

\begin{table*}[t]\vspace{-0.3cm}
  \renewcommand{\arraystretch}{1.}
  \renewcommand{\tabcolsep}{5.pt}
    \centering
    \caption{Metrics of using (ours) and not using (baselines) the dependency regularization. All figures are mean of five independent runs.}
    \label{tab:robustness}
    \vspace{-0.3cm}
    \begin{tabular}{l||ccc|ccc|ccc}
    \toprule
    % \Xhline{2.\arrayrulewidth}
    \multicolumn{1}{c||}{} & \multicolumn{3}{c|}{\textbf{ImageNet Acc. {$(\uparrow)$}}} & \multicolumn{3}{c|}{\textbf{Dependency Coefficients} {$(\downarrow)$}} & \multicolumn{3}{c}{\textbf{ImageNet-O AUPR} {$(\uparrow)$}}\\ \cline{2-4} \cline{5-7} \cline{8-10}
    \multicolumn{1}{c||}{\multirow{-2}{*}{\textbf{Backbone}}} & \textit{Baseline} & \textit{Ours} & \textit{Impro} & \textit{Baseline} & \textit{Ours} & \textit{Impro} & \textit{Baseline} & \textit{Ours} & \textit{Impro}\\ 
    \hline
    % \Xhline{2.\arrayrulewidth}
      ResNet18   & 69.83  & 70.12 & +0.29 & 0.70  & 0.02 & +0.68 & 15.15  & 15.48 & +0.33\\
    ResNet50 & 76.37 & 76.66 & +0.29 & 1.10  & $4.5e^{-4}$ & +1.10 & 13.98  & 14.07 & +0.09 \\ 
    \hline
       Vit-S   & 80.67 & 81.52 & +0.85 & 0.1  & $3.1e^{-3}$ & +0.1 & 28.54  & 31.14 & +2.60  \\
       Swin-T & 82.16 & 82.18 & +0.02 & 0.39  & 0.01 & +0.38 & 27.66  & 28.13 & +0.47 \\ 
    \bottomrule
    % \Xhline{2.\arrayrulewidth}
    \end{tabular}
    % \vspace{-0.1in}
\end{table*}
 
\begin{table}[ht]\vspace{-0.4cm}
  \renewcommand{\arraystretch}{1.}
  \renewcommand{\tabcolsep}{5.pt}
    \centering
    \caption{Classification accuracy in base (900) and new (100) categories separately. While much simpler, learning new categories through neural dependencies outperform baselines if only consider the performance in the new categories. All figures are mean of five independent runs.}
    \label{tab:generalization_3}\vspace{-0.3cm}
    \renewcommand{\arraystretch}{1.05}
    \begin{tabular}{l||cc|cc}
    \toprule
    % \Xhline{2.\arrayrulewidth}
    \multicolumn{1}{c||}{} & \multicolumn{2}{c|}{ResNet-50} & \multicolumn{2}{c}{Swin-T} \\ \cline{2-5} 
    \multicolumn{1}{c||}{\multirow{-2}{*}{\textbf{Method}}} & \textit{900} & \textit{100} & \textit{900} & \textit{100}\\ 
    \hline
    \textit{Baseline}   &72.54 &36.94 &71.42 & 72.04   \\ 
    \hline
    \textit{Ours} &72.89 &48.90 &71.36 &74.63   \\ 
    \bottomrule
    % \Xhline{2.\arrayrulewidth}
    \end{tabular}
    % \vspace{-0.1in}
\end{table}

\newcommand{\vTheta}{\bm{\Theta}}

\subsection{Generalizability}\label{subsec:generalizability}
Now that the logits of one category can be well predicted by the logits of some others, we are curious about whether we can learn a cluster of base categories, and then predict new classes purely using linear combinations of the logits of those base categories. Especially, can the overall performance of this setting be comparable to training the baseline model on the whole dataset? This problem is of general interest to many machine learning scenarios. 1) \textit{Incremental Learning.} In incremental learning~\cite{wu2019large,mittal2021essentials,castro2018end} we need to learn to predict novel categories using a pretrained network on old categories. Typical methods will finetune the pretrained network in the new categories to achieve this goal, which then arouses concerns of damaging the knowledge of the old categories. Using our setting we can explore the potential of keeping the pretrained network unchanged and learning merely a small weight matrix to handle novel categories, which is cheap and efficient to train and deploy in various devices and realistic scenarios. 2) \textit{Transfer Learning.} A similar but different occasion is transfer learning~\cite{pan2009survey,weiss2016survey,wang2018deep}, where we seek to take advantage of knowledge of old domains to improve performance in new data domains. While categories are also instances of domains, our setting also explores a new way of knowledge transfer among domains.  3) \textit{Representation Learning.} Our setting can partially reveal how representations~\cite{bengio2013representation} of base knowledge help classifications in out-of-distribution data (new categories). Future studying of this setting may reveal the source of the generalizability of neural networks from the perspective of neural dependencies.

To implement our setting, we may first train a deep classification network $f_{\rm base}:\sR^m\rightarrow\sR^{n_1}$ on the $n_1$ base categories. Then we learn a coefficient matrix $\vTheta\in\sR^{n_1\times n_2}$ by fixing the parameters of $f_{\rm base}$ and minimizing the training loss of $f_{\rm new}=f_{\rm base}\vTheta$ on the training set of the new categories. We then concatenate $f_{\rm all}=[f_{\rm base}, f_{\rm base}\vTheta]^T$ to form a new classifier for all the categories.
We sample 500 samples per category from the trainset of ImageNet-1k as our training data; the remains are used for constructing a balanced binary testing set we will use later. We evaluate the following three settings: 1) from 900 base classes to 100 new classes ($900\to100$), 2) from 950 base classes to 50 new classes ($950\to50$), and 3) from 999 base classes to 1 new class ($999\to1$) within a dataset. To approach a binary classification scenario, for $999\to1$ case we additionally test on 500 positive and negative sample pairs from the 
remained trainset of ImageNet as the \textbf{\textbf{$999\to1(pos\&neg)$}} setting. The baselines $f_{\rm baseline}$ are backbone models trained on the whole 1,000 category training data. Other details can be found in appendix.

\noindent\textbf{Experimental Results.} 
We report the performance of $f_{\rm all}$ and $f_{\rm baseline}$  in~\cref{tab:generalization}, where we can find both settings (ours v.s. baselines) achieve comparable performance. While our setting requires training on only a small coefficient matrix, it consumes much less computation and time resources (less than 60\% time consumption of the baseline in each epoch, see appendix for detail) compared with the baselines. We further investigate how our setting performs in the new categories. \cref{tab:generalization_3} reports classification accuracy in the old 900 and new 100 categories of our setting and baselines (here we choose the class with maximum logits in the 900/100 categories as the prediction results). We can find that our setting significantly outperforms the baselines in the new classes. Both results reveal the power of neural dependencies in the generalizability of deep networks.

%From the results shown in~\cref{tab:generalization}, we observed that our method achieves comparable results in terms of all settings. Specifically, our method achieves a performance close to the overall model training with only the classifier trained. In the \textbf{\textbf{$999\to1(pos\&neg)$}} setting, our method underperforms baseline by 0.8\%/1.8\% in the ResNet and Swin-T, respectively. Nonetheless, with a lot of training time saved, our method still can achieve comparable results. 
%To better evaluate the potentials of neural dependencies in general scenarios, we train a classifier for the new 100 categories based on the well-trained model of 900 classes. As shown in~\cref{tab:generalization_3}, the results depict the relative improvement of our proposed strategy in the new 100 classes when compared to the base models.
%These results demonstrate that the logits of some categories can be directly obtained by linearly combining the predictions of other categories.
  
\subsection{Robustness}\label{subsec:robustness}
As we have mentioned before, some neural dependencies are not that sensible for humans. We are therefore curious about whether cutting off those dependencies can help the network and improve robustness. Here we compare two cases, the baselines and baselines finetuned by adding the regularization term $\vert \E_{\vx\sim\pdata}[f(\vx)_if(\vx)_j]\vert$ where $c_i,c_j$ are the two categories that emerge irrational neural dependencies to cut off. We use two benchmarks, ImageNet-1k and ImageNet-O~\cite{hendrycks2021natural}. ImageNet-O consists of images from 200 classes that are unseen in ImageNet-1k, and is used to test the robustness of networks to out-of-distribution samples. This ability is usually measured by the AUPR (\textit{i.e., area under the precision-recall curve}) metric~\cite{boyd2013area}. This metric requires anomaly scores, which is the negative of the maximum softmax probabilities from a model that can classify the 200 classes. We train the baseline models for 90 epochs and our settings for 60 epochs of regular training followed by 30 epochs of finetuning using the regularization term $\vert \E_{\vx\sim\pdata}[f(\vx)_if(\vx)_j]\vert$. We manually choose one dependency to cut off for each case. Details can be found in appendix.

% \noindent\textbf{Dataset and Settings.} 
% To evaluate the effect of neural dependencies on the model robustness, we conduct comprehensive experiments on two benchmarks ImageNet-1k and ImageNet-O. ImageNet-1k contains 1000 classes, providing a large-scale evaluation scenery with average classification accuracy. ImageNet-O \cite{hendrycks2021natural} consists of images from 200 classes that are not found in the ImageNet-1k dataset. It is used to test the robustness of vision models to out-of-distribution samples, which is reported using the AUPR (\textit{i.e., area under the precision-recall curve}) metric. (AUPR). This metric requires anomaly scores, which is the negative of the maximum softmax probabilities from a model that can classify the 200 classes.
% For a comprehensive analysis, we conduct the comparative experiments on four standard backbones ResNet18 \cite{he2016deep}, ResNet50, ViT-S \cite{dosovitskiy2020image} and Swin-T \cite{liu2021swin}. For a fair comparison, we finetune the pretrained models with the same epochs (\textit{i.e., 30 epochs}) given the corresponding limits. To reduce the error caused by random factors, we select different random seeds to test 5 times and then take the average for each model. We use a SGD optimizer, in which the learning rate is set to 0.01. The attenuation rate is set to 0.0005, and batch size is set to 128. 

\noindent\textbf{Experimental Results.} 
% \todo{robustness}
\cref{tab:robustness} reports the results. The regularization term does cut off the neural dependencies as the dependency coefficients are approaching zero after regularization. This then results in some improvements of performance in both ImageNet and ImageNet-O for all the backbones. While here we only cut-off one dependency for each case, we believe a thorough consideration of reasonable dependencies to maintain may benefit the network more. This reveals the connection between neural dependencies and the robustness of networks.

%To demonstrate the function of neural dependencies, we valuate two finetuned models on three metrics. For the baseline model, we simply finetune the pretrained models without any limits. For our models, we introduce the dependency regularization on the output logits, which is calculated by estimating the first order norm of class-specific correlations. We select different regularized classes for different backbones, which are consistent with the classes in~\cref{fig:ND_exps}. The average classification accuracy on ImageNet-1k, the lasso values on ImageNet-1k and the AUPR values on ImageNet-O are reported in~\cref{tab:robustness}. Firstly, it can be shown that the removal of the negative neural dependencies promotes the overall discrimination in all backbones. Then, for the regularized classes in each backbone, the corresponding lasso values are obviously reduced, mitigating the improper correlations in the initial models. Finally, the AUPR values on Imagenet-O dataset are improved especially on larger models. It demonstrates that the consideration on the neural dependencies helps to improve the overall robustness of the network, especially when dealing with adversarial samples.
\section{Conclusion}\label{sec:conclusion}
This paper reveals an astonishing neural dependency phenomenon emerging from learning massive categories. Given a well-trained model, the logits predicted for some category can be directly obtained by linearly combining the predictions of a few others. Theoretical investments demonstrate how to find those neural dependencies precisely, when they happen, and why the dependency is usually sparse, \textit{i.e.} only a few instead of numerous of other categories related to one target category. Further empirical studies reveal multiple attractive potentials of neural dependencies from the aspects of visualization, generalization, and robustness of deep classification networks.
{\small
\bibliographystyle{ieee_fullname}
\bibliography{ref}
}

\appendix
\renewcommand\thefigure{A\arabic{figure}}
\renewcommand\thetable{A\arabic{table}}  
\renewcommand\theequation{A\arabic{equation}}
\renewcommand\thealgorithm{A\arabic{algorithm}}

\clearpage
\onecolumn
\section{Proof}
\subsection{Proof to \cref{th:equi}}\label{pf:equi}
This is the natural results of \cref{eq:markov ineq}.

\subsection{Property of Function $\bm{err}_i(\vtheta^*(\lambda))$}\label{pf:err_path}
The continuity of the solution path and the existence of $\lammax=2\Vert\Cov_{[n]\setminus i}^i\Vert_{\infty}$ are natural results of the property of general Lasso regressions \cite{tibshirani1996regression,tibshirani2011solution}. Here we prove that
\begin{gather}\label{eq:err_path1}
    \bm{err}_i(\vtheta^*(0)=\frac{\det[\Cov]}{\det[\Cov_{[n]\setminus i}^{[n]\setminus i}]}, \\
    \bm{err}_i(\vtheta^*(\lambda_{\rm{max}}))=\Cov_i^i.\label{eq:err_path2}
\end{gather}
In fact, the second equality \cref{eq:err_path2} is easy to verify directly, so we only need to prove the first one, \cref{eq:err_path1}. Let 
\begin{gather}
    \Cov=\bm{U}\bm{\Gamma}\bm{U}^T,\\
\bm{U}\bm{U}^T=\bm{I},\\
\bm{\Gamma}={\rm{diag}}\{\gamma_1^2,\cdots,\gamma_n^2\},\\
\vtheta=\sum_{j=1}^n\bm{\alpha}_j\bm{U}^j,\\
\vtheta_i=\sum_{j=1}^n\bm{\alpha}_j\bm{U}^j_i=\bm{U}_i\bm{\alpha}=-1,
\end{gather}
and $\bm{U}^1,\cdots,\bm{U}^n$ be the eigenvectors of $\Cov$. The original problem \cref{problem:cLasso} (when $\lambda=0$) now becomes
%\begin{equation}
%    \min_{\bm{\alpha}}\sum_{j=1}^n\bm{\alpha}_j^2\gamma_j^2,\\
%\text{subject to }\sum_{j=1}^n\bm{\alpha}_j\bm{U}^j_i=-1.
%\end{equation}
%
\begin{equation}
\begin{aligned}
    &\min_{\bm{\alpha}}\sum_{j=1}^n\bm{\alpha}_j^2\gamma_j^2,\\
&~\text{subject to }\sum_{j=1}^n\bm{\alpha}_j\bm{U}^j_i=-1.
\end{aligned}
\end{equation}
Using Lagrange multiplier, the following problem will attain extreme value together with the above problem
\begin{equation}
    \min_{\bm{\alpha},\eta}H(\bm{\alpha},\eta)=\sum_{j=1}^n\bm{\alpha}_j^2\gamma_j^2+\eta(\sum_{j=1}^n\bm{\alpha}_j\bm{U}^j_i+1).
\end{equation}
Thus we have
\begin{gather}
    \frac{\partial H}{\partial\bm{\alpha}_j}=2\bm{\alpha}_j\gamma_j^2+\eta\bm{U}_i^j=0\Leftrightarrow\bm{\alpha}_j=-\frac{\eta\bm{U}_i^j}{2\gamma_j^2},j=1,\cdots,n,\\
\Leftrightarrow \bm{\alpha}=-\frac{\eta}{2}\bm{\Gamma}^{-1}\bm{U}_i^T=-\frac{\eta}{2}\bm{\Gamma}^{-1}\bm{U}^T\bm{e}_i \\
\frac{\partial H}{\partial\eta} =\sum_{j=1}^n\bm{\alpha}_j\bm{U}^j_i+1=0\Leftrightarrow \sum_{j=1}^n-\frac{\eta\bm{U}_i^j}{2\gamma_j^2}\bm{U}_i^j=-1\Leftrightarrow\frac{\eta}{2}\sum_{j=1}^n\frac{(\bm{U}_i^j)^2}{\gamma_j^2}=1\label{eq:eta1}\\
\Leftrightarrow \eta=2(\bm{U}_i\bm{\Gamma}^{-1}\bm{U}_i^T)^{-1}=2(\bm{e}_i^T\bm{U}\bm{\Gamma}^{-1}\bm{U}^T\bm{e}_i)^{-1}=2/(\Cov^{-1})_i^i.\label{eq:eta2}
\end{gather}
Combining the above derivation, we have
\begin{gather}
    \vtheta^*(0)=\bm{U}\bm{\alpha}^*=-\frac{\eta}{2}\bm{U}\bm{\Gamma}^{-1}\bm{U}^T\bm{e}_i=-\frac{\eta}{2} \Cov^{-1}\bm{e}_i,\\
(\vtheta^*(0)^T\Cov\vtheta^*(0))=\frac{\eta^2}{4}\bm{e}_i^T\Cov^{-T}\Cov\Cov^{-1}\bm{e}_i=\frac{\eta^2}{4}\bm{e}_i^T\Cov^{-1}\bm{e}_i\\
=\frac{\eta^2}{4}(\Cov^{-1})_i^i=1/(\Cov^{-1})_i^i=\frac{\det[\Cov]}{\det[\Cov_{[n]\setminus i}^{[n]\setminus i}]}.
\end{gather}

\subsection{Case of Small $\Cov_i^i$}\label{pf:small_cov}
Here we may want the ratio
\begin{equation}
    \gR(\Cov,i)=\frac{\det[\Cov]}{\Cov_i^i\det[\Cov_{[n]\setminus i}^{[n]\setminus i}]}=\frac{1}{\sum_{j=1}^n\frac{\valpha_j^2}{\sigma_j^2}\sum_{j=1}^n\valpha_j^2\sigma_j^2}
\end{equation}
to be as small as possible, where $\sigma_1^2\geq\cdots\geq\sigma_n^2$ are the eigenvalues of $\Cov$ and $\vq_1,\cdots,\vq_n$ are the corresponding eigenvectors, $\alpha_j=\langle\ve_i,\vq_j\rangle,j\in[n]$ (refer to deduction in Appendix). This is also the minimum relative prediction error, \ieno,
\begin{equation}
    \inf_{\lambda\geq 0}\frac{\E_{\vx\sim\pdata}[\vert f(\vx)_i-\sum_{j\neq i}\vtheta^*(\lambda)_j f(\vx)_j\vert^2]}{\E_{\vx\sim\pdata}[\vert f(\vx)_i\vert^2]}= \gR(\Cov,i),\forall\lambda\in[0,\lammax].
\end{equation}
Geometrically, this means that the $i$-th coordinate axis admits valid components in the eigenvectors of both non-tiny and tiny eigenvalues.

\subsection{Property of the Lower Bound $\bm{err}_i(\vtheta^*(0))$}\label{pf:lower_bound}
Here we prove that
\begin{equation}
    \bm{err}_i(\vtheta^*(0))=\frac{\det[\Cov]}{\det[\Cov_{[n]\setminus i}^{[n]\setminus i}]}=\left(\sum_{j=1}^n\frac{\valpha_j^2}{\sigma_j^2}\right)^{-1}.
\end{equation}
This is the natural result of \cref{eq:eta1,eq:eta2}, as
\begin{equation}
    1/(\Cov^{-1})_i^i=\frac{\det[\Cov]}{\det[\Cov_{[n]\setminus i}^{[n]\setminus i}]}=1/(\Cov^{-1})_i^i=\frac{\eta}{2}=\left(\sum_{j=1}^n\frac{(\bm{U}_i^j)^2}{\gamma_j^2}\right)^{-1}.
\end{equation}
Let $\mU^j=\vq_j$ and $\sigma_j^2=\gamma_j^2$. Then we obtain the result.

\subsection{Sparsity Condition of the Solution}\label{pf:sparse_cond}
By KKT conditions, the optimal value of \cref{problem:classo2} is attained only if
\begin{equation}
    \vzero\in\hCov\hvtheta^*(\lambda)-\hvb+\frac{\lambda}{2}\partial\Vert\hvtheta\Vert_1
\end{equation}
where $\partial\Vert\hvtheta\Vert_1=\{\vv:\Vert\vv\Vert_{\infty}\leq 1, \vv^T\hvtheta=\Vert\hvtheta\Vert_1\}$ is the subgradient of $\Vert\cdot\Vert_1$. By Cauchy inequality,
\begin{equation}
    \Vert\hvtheta\Vert_1=\vv^T\hvtheta\leq\Vert\vv\Vert_{\infty}\Vert\hvtheta\Vert_1=\Vert\hvtheta\Vert_1.
\end{equation}
The equality holds if and only if
\begin{equation}
    \vert\vv_i\vert<1\Rightarrow \hvtheta_i=0.
\end{equation}
Thus we have the sparsity condition of the solution.

\subsection{Proof to \cref{th:slope}}\label{pf:slope}
To start, we deduce the dual problem of \cref{problem:cLasso}. For standard Lasso problem
\begin{equation}
    \min_{\bm{\beta}}\frac{1}{2}\Vert\bm{y}-\bm{X}\bm{\beta}\Vert_2^2+\lambda\Vert\bm{\beta}\Vert_1,
\end{equation}
where $\vy$ are labels and $\mX$ are observations, its dual problem is~\cite{tibshirani2011solution}
\begin{equation}
\begin{aligned}
    &\max_{\bm{\xi}}\frac{1}{2}\Vert\bm{y}\Vert_2^2-\frac{\lambda^2}{2}\Vert\bm{\xi}-\frac{\bm{y}}{\lambda}\Vert_2,\\
    &\text{ subject to }\vert(\bm{X}^j)^T\bm{\xi}\vert\leq1,j=1,\cdots,n.
\end{aligned}
\end{equation}
Let 
\begin{gather}
\hCov=\bm{Q}\bm{\Sigma}\bm{Q}^T,\\
\bm{A}=\bm{Q}\bm{\Sigma}^{1/2}\bm{Q}^T,\\
\hvb=\Cov_{[n]\setminus i}^i,\\
\bm{y}=\sqrt{2}\bm{A}^{-T}\hvb,\\
\text{and}~\bm{X}=\sqrt{2}\bm{A}.
\end{gather}
We then get the dual problem of \cref{problem:cLasso,problem:classo2} as
\begin{equation}\label{problem:dual}
\begin{aligned}
    &\max_{\bm{\xi}} \Vert\bm{A}^{-1}\hvb\Vert_2^2-\frac{\lambda^2}{2}\Vert\bm{\xi}-\frac{\sqrt{2}\bm{A}^{-T}\hvb}{\lambda}\Vert_2^2,\\
    &\text{ subject to }\Vert\bm{A}\bm{\xi}\Vert_{\infty}\leq \frac{\sqrt{2}}{2}.
\end{aligned}
\end{equation}
By the KKT condition, we further have
\begin{gather}\label{eq:dual_kkt}
    \sqrt{2}\bm{A}^{-T}\hvb=\sqrt{2}\bm{A}\hvtheta^*(\lambda)+\lambda\bm{\xi}^*(\lambda),\\
    \text{when }\lambda\geq\lambda_{\rm{max}}=\Vert\sqrt{2}\bm{A}\sqrt{2}\bm{A}^{-T}\hvb\Vert_{\infty}=2\Vert \hvb\Vert_{\infty},\hvtheta^*=\bm{0}.
\end{gather}

The dual problem \cref{problem:dual} can be further transferred into
\begin{equation}
\begin{aligned}
    &\min_{\bm{\xi}} \Vert\bm{\xi}-\frac{\sqrt{2}\bm{A}^{-T}\hvb}{\lambda}\Vert_2^2,\\
    &\text{ subject to }\Vert\bm{A}\bm{\xi}\Vert_{\infty}\leq \frac{\sqrt{2}}{2}.
\end{aligned}
\end{equation}
\newcommand{\vxi}{\bm{\xi}}
This problem solves the projection of point $\frac{\sqrt{2}\bm{A}^{-T}\hvb}{\lambda}$ onto the convex set $\{\vxi:\Vert\bm{A}\bm{\xi}\Vert_{\infty}\leq \frac{\sqrt{2}}{2}\}$. Denote its solution as $\vxi^*(\lambda)$ for parameter $\lambda$. It is then easy to verify 
\begin{gather}\label{eq:slope}
\Vert\frac{\sqrt{2}\mA^{-T}\hvb}{\lambda''}-\frac{\sqrt{2}\mA^{-T}\hvb}{\lambda'}\Vert_2^2=\Vert\bm{\xi^*}(\lambda'')-\bm{\xi^*}(\lambda')-\vxi^*(\lambda'')+\frac{\sqrt{2}\mA^{-T}\hvb}{\lambda''}+\vxi^*(\lambda')-\frac{\sqrt{2}\mA^{-T}\hvb}{\lambda'}\Vert_2^2\\
    =\Vert\bm{\xi^*}(\lambda'')-\bm{\xi^*}(\lambda')\Vert_2^2+\Vert\vxi^*(\lambda'')-\frac{\sqrt{2}\mA^{-T}\hvb}{\lambda''}\Vert_2^2+ \Vert\vxi^*(\lambda')-\frac{\sqrt{2}\mA^{-T}\hvb}{\lambda'}\Vert_2^2\\
    +2\langle \vxi^*(\lambda'')-\vxi^*(\lambda'),\frac{\sqrt{2}\mA^{-T}\hvb}{\lambda''}-\vxi^*(\lambda'')\rangle+2\langle \vxi^*(\lambda'')-\vxi^*(\lambda'),\vxi^*(\lambda')-\frac{\sqrt{2}\mA^{-T}\hvb}{\lambda'}\rangle\\
    +2\langle\frac{\sqrt{2}\mA^{-T}\hvb}{\lambda''}-\vxi^*(\lambda''),\vxi^*(\lambda')-\frac{\sqrt{2}\mA^{-T}\hvb}{\lambda'}\rangle \\
    =\Vert\bm{\xi^*}(\lambda'')-\bm{\xi^*}(\lambda')\Vert_2^2+\Vert\vxi^*(\lambda'')-\frac{\sqrt{2}\mA^{-T}\hvb}{\lambda''}-\vxi^*(\lambda')-\frac{\sqrt{2}\mA^{-T}\hvb}{\lambda'}\Vert_2^2\\
    +2\langle \vxi^*(\lambda'')-\vxi^*(\lambda'),\frac{\sqrt{2}\mA^{-T}\hvb}{\lambda''}-\vxi^*(\lambda'')\rangle+2\langle \vxi^*(\lambda'')-\vxi^*(\lambda'),\vxi^*(\lambda')-\frac{\sqrt{2}\mA^{-T}\hvb}{\lambda'}\rangle\\
    \geq \Vert\bm{\xi^*}(\lambda'')-\bm{\xi^*}(\lambda')\Vert_2^2.
\end{gather}
The last inequality uses the fact that
\begin{gather}
    2\langle \vxi^*(\lambda'')-\vxi^*(\lambda'),\frac{\sqrt{2}\mA^{-T}\hvb}{\lambda''}-\vxi^*(\lambda'')\rangle\geq0,\\
    2\langle \vxi^*(\lambda'')-\vxi^*(\lambda'),\vxi^*(\lambda')-\frac{\sqrt{2}\mA^{-T}\hvb}{\lambda'}\rangle\geq0,
\end{gather}
for convex set $\{\vxi:\Vert\bm{A}\bm{\xi}\Vert_{\infty}\leq \frac{\sqrt{2}}{2}\}$ and the projections $\vxi^*(\lambda'),\vxi^*(\lambda'')$ on it. Thus we have
\begin{equation}\label{eq:slop_proof}
    \Vert\bm{\xi^*}(\lambda'')-\bm{\xi^*}(\lambda')\Vert_2\leq \Vert\frac{\sqrt{2}\mA^{-T}\hvb}{\lambda''}-\frac{\sqrt{2}\mA^{-T}\hvb}{\lambda'}\Vert_2.
\end{equation}
Combining \cref{eq:dual_kkt}, we then get the result of this theorem.

\subsection{Finer Estimation of the Value of $\bm{err}_i(\vtheta^*(\lambda))$}
We have
\begin{gather}
    \vert\vtheta^*(\lambda'')^T\Cov\vtheta^*(\lambda'')-\vtheta^*(\lambda')^T\Cov\vtheta^*(\lambda')\vert\\
    =\vert\hat{\vtheta}^*(\lambda'')^T\hat{\Cov}\hat{\vtheta}^*(\lambda'')-\hat{\vtheta}^*(\lambda')^T\hat{\Cov}\hat{\vtheta}^*(\lambda')-2\hvb^T(\hat{\vtheta}^*(\lambda'')-\hat{\vtheta}^*(\lambda'))\vert\\
    =\vert\Vert\bm{A}\hat{\vtheta}^*(\lambda'')-\bm{A}^{-T}\hvb\Vert_2^2-\Vert\bm{A}\hat{\vtheta}^*(\lambda')-\bm{A}^{-T}\hvb\Vert_2^2\vert\\
    =\vert\frac{\lambda''^2}{2}\Vert\bm{\xi}^*(\lambda'')\Vert_2^2-\frac{\lambda'^2}{2}\Vert\bm{\xi}^*(\lambda')\Vert_2^2\vert.
\end{gather}
Setting $\lambda'=\lambda_{\rm{max}},\lambda''=\lambda$, we can have 
\begin{gather}
    0\leq\Cov_i^i-\vtheta^*(\lambda)^T\Cov\vtheta^*(\lambda)=\Vert\bm{A}^{-T}\hvb\Vert_2^2-\frac{\lambda^2}{2}\Vert\bm{\xi}^*(\lambda)\Vert_2^2\\
\leq \Vert\bm{A}^{-T}\hvb\Vert_2^2-\frac{\lambda^2}{2}(\Vert\frac{\sqrt{2}}{2\Vert\hvb\Vert_{\infty}}\bm{A}^{-T}\hvb\Vert_2+\Vert\bm{\xi}^*(\lambda)-\bm{\xi}^*(\lambda_{\rm{max}})\Vert_2)^2\\
\leq \Vert\bm{A}^{-T}\hvb\Vert_2^2-\frac{\lambda^2}{2}(\Vert\frac{\sqrt{2}}{2\Vert\hvb\Vert_{\infty}}\bm{A}^{-T}\hvb\Vert_2+\Vert\bm{\xi}^*(\lambda)-\bm{\xi}^*(\lambda_{\rm{max}})\Vert_2)^2.
\end{gather}
Taking \cref{eq:slop_proof} into the above result yields finer estimation to the value of $\bm{err}_i(\vtheta^*(\lambda))=\vtheta^*(\lambda)^T\Cov\vtheta^*(\lambda)$.

\subsection{Proof to \cref{th:screening}}\label{pf:screening}
This is the natural result of \cref{eq:pre_screen}.

\section{Experiment Setting}
\noindent\textbf{Experiment Setup in \cref{sec:method}.} We use the official pretrained models for all the experiments in this section. For ResNets, we use the official Pytorch pretrained models\footnote{\url{https://github.com/pytorch/examples/tree/main/imagenet}}. For VIT, we use the official checkpoints provided by Google Research\footnote{\url{https://github.com/google-research/vision_transformer}}. For Swin-T, we use the official pretrained model provided by Microsoft\footnote{\url{https://github.com/microsoft/Swin-Transformer}}.

\noindent\textbf{Lasso Solver.}
We use the ${\rm sklearn.linear\_model.Lasso}$ of sklearn~\cite{scikit-learn} package to solve the CovLasso regression in this paper. ${\rm max\_iter}$ is set to 50,000, ${\rm alpha}$ is set to 0.25 for Swin-T and 2.5 for the remaining algorithms. All the other hyper-parameters are set as default.

\noindent\textbf{Training settings of \cref{subsec:generalizability}.} 
For both ResNet-50 and Swin-T, following the conventional setting, we first perform intermediate pre-training of a ResNet $f_{\rm base}:\sR^m\rightarrow\sR^{n_1}$ on the $n_1$ base categories of ImageNet1K for 90 epochs with image resolution 224$\times224$. Then we learn a coefficient matrix $\vTheta\in\sR^{n_1\times n_2}$ by fixing the parameters of $f_{\rm base}$ and training on the training set of the new categories for 60 epochs. 
For ResNet-50, we use SGD with mini-batch size 256 on 8 Nvidia-A100 GPUs. The learning rate starts from 0.1 and is divided by 10 on the 30-th and 60-th epoch, and we use a weight decay of 0.0001 and a momentum of 0.9.
For Swin-T, we use AdamW with a mini-batch size of 256 on 8 A100 GPUs. The learning rate starts from 0.002
and is divided by 10 on the 60-th and 80-th epochs, and we use a weight decay of 0.05.

\noindent\textbf{Training settings of \cref{subsec:robustness}.} 
During the fine-tuning process of all backbones, we use an SGD optimizer, in which the initial learning rate is set to 0.01 for 30 epochs. We use a weight decay of 0.0005 and a momentum of 0.9. The batch size is set to 256. The loss weight for the regularization term is set to 0.2, and eight NVIDIA Tesla A100 GPUs are used for all experiments. All datasets adopted in this paper are open to the public.

\section{More Results}
\begin{figure}[t]
\centering
\includegraphics[width=0.98\linewidth]{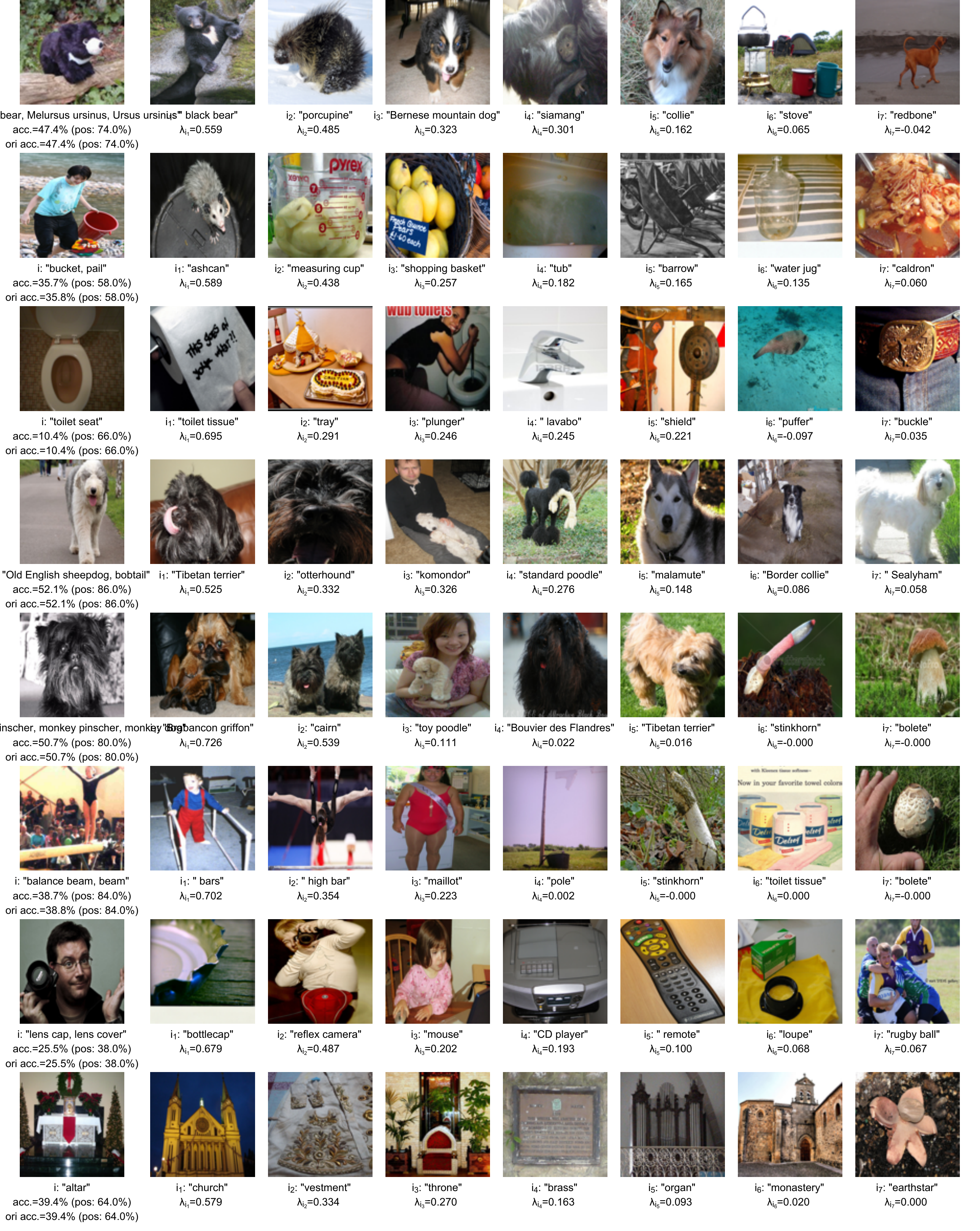}
\caption{Results from ResNet-18, where `acc.' and `ori acc.' denote the classification accuracies on the ImageNet validation set, while `pos: xx\%' is the accuracy on positive samples only.}
\label{fig:supp_deficit_res18}
\end{figure}

\begin{figure}[t]
\centering
\includegraphics[width=0.98\linewidth]{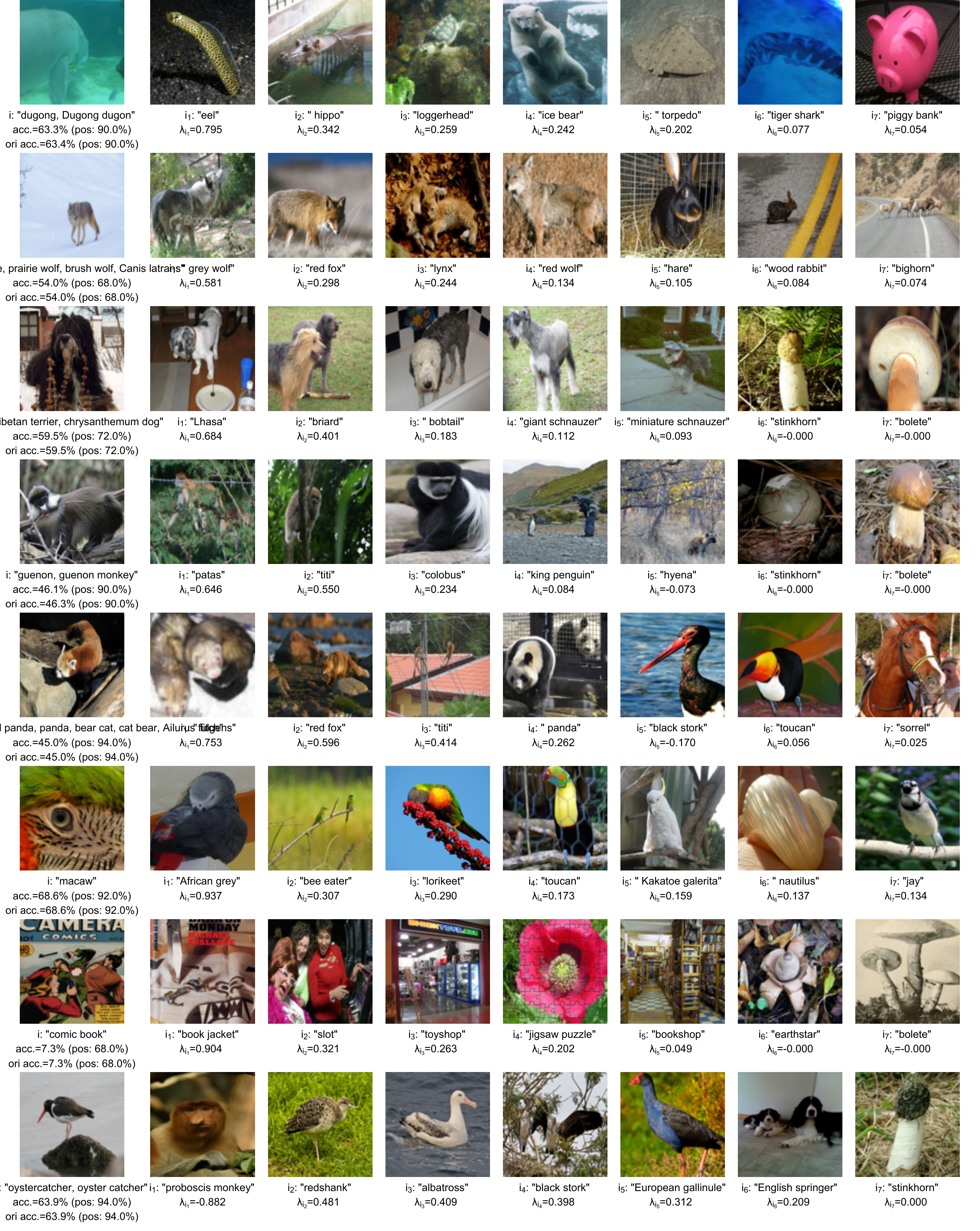}
\caption{Results from ResNet-50, where `acc.' and `ori acc.' denote the classification accuracies on the ImageNet validation set, while `pos: xx\%' is the accuracy on positive samples only.}\label{fig:supp_deficit_res50}
\end{figure}

\begin{figure}[t]
\centering
\includegraphics[width=0.98\linewidth]{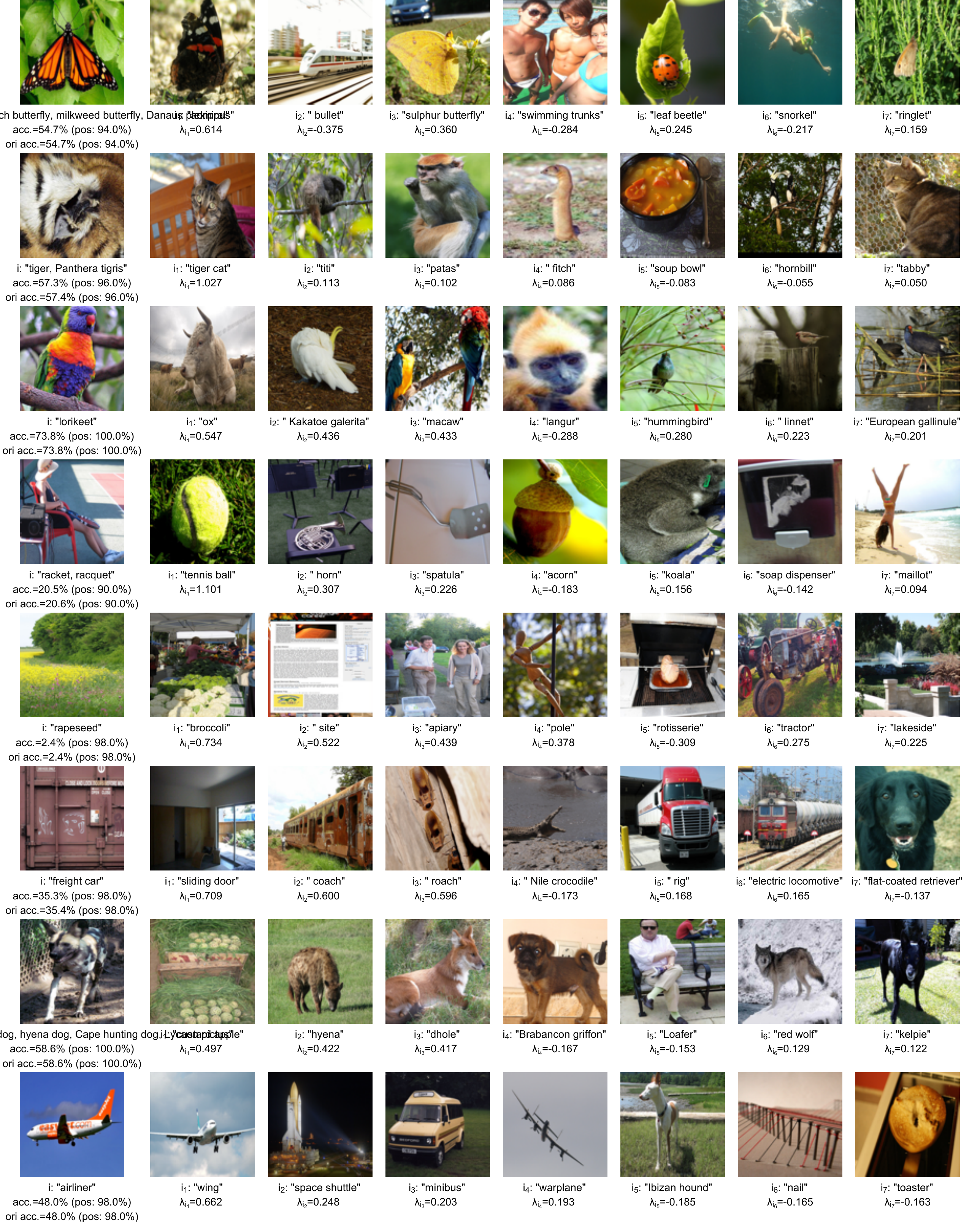}
\caption{Results from ViT-S, where `acc.' and `ori acc.' denote the classification accuracies on the ImageNet validation set, while `pos: xx\%' is the accuracy on positive samples only.}\label{fig:supp_deficit_vit}
\end{figure}

\begin{figure}[t]n
\centering
\includegraphics[width=0.98\linewidth]{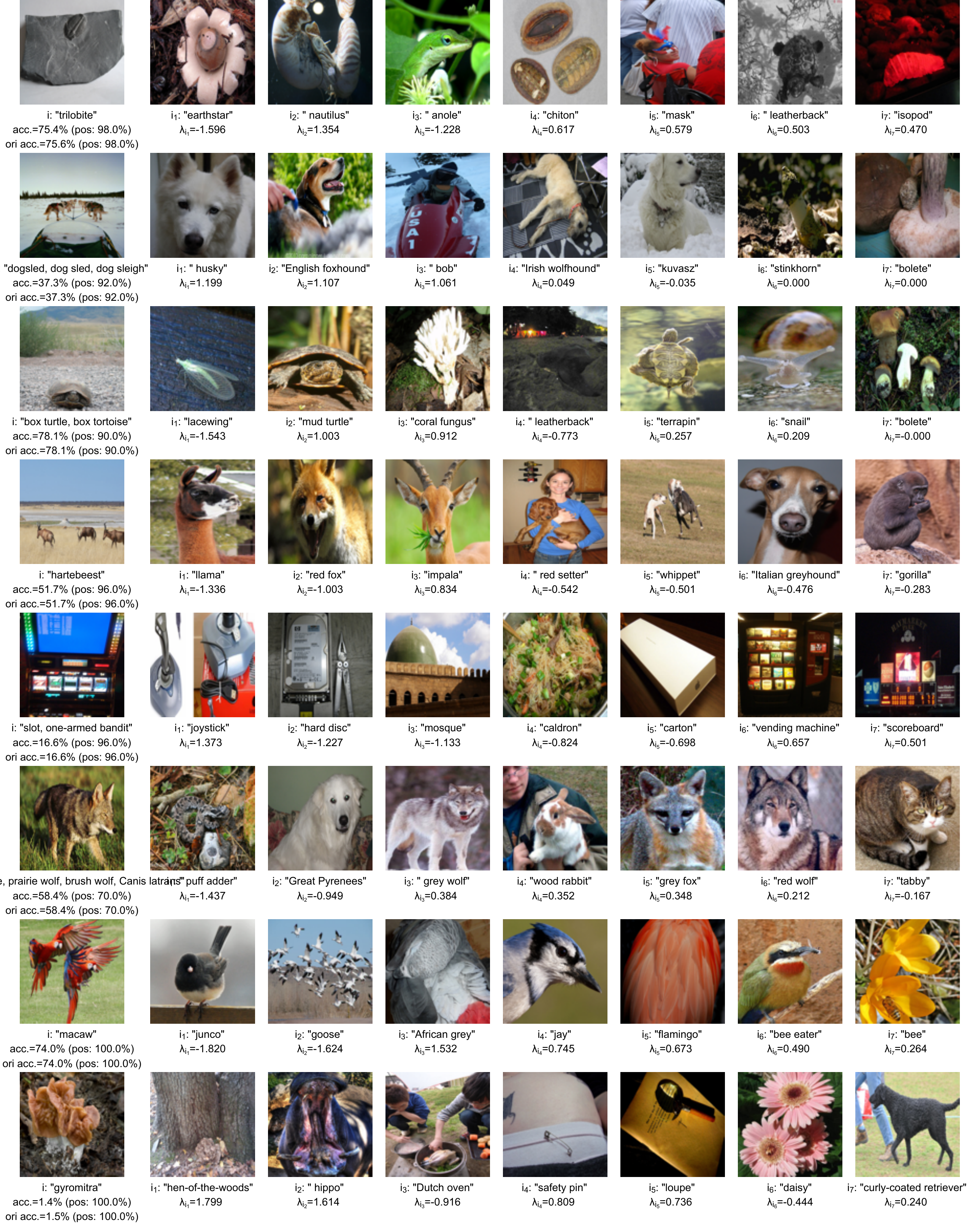}
\caption{Results from Swin-T, where `acc.' and `ori acc.' denote the classification accuracies on the ImageNet validation set, while `pos: xx\%' is the accuracy on positive samples only.}\label{fig:supp_deficit_swin}
\end{figure}

\begin{figure}[t]
\centering
\includegraphics[width=0.98\linewidth]{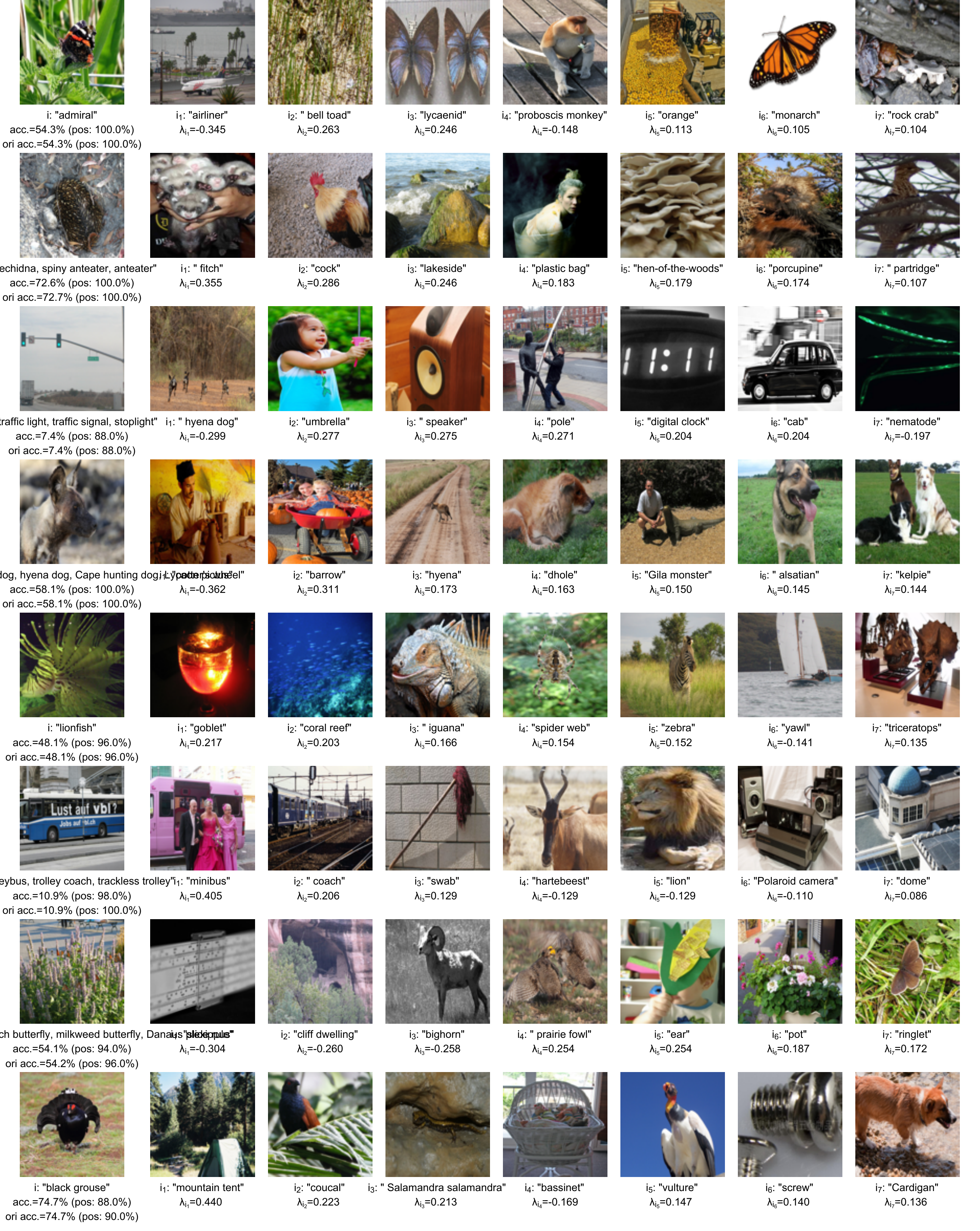}
\caption{Results from ResNet-50$\to$Swin-T, where `acc.' and `ori acc.' denote the classification accuracies on the ImageNet validation set, while `pos: xx\%' is the accuracy on positive samples only.}\label{fig:supp_deficit_resnet50_swim_t}
\end{figure}

\begin{figure}[t]
\centering
\includegraphics[width=0.98\linewidth]{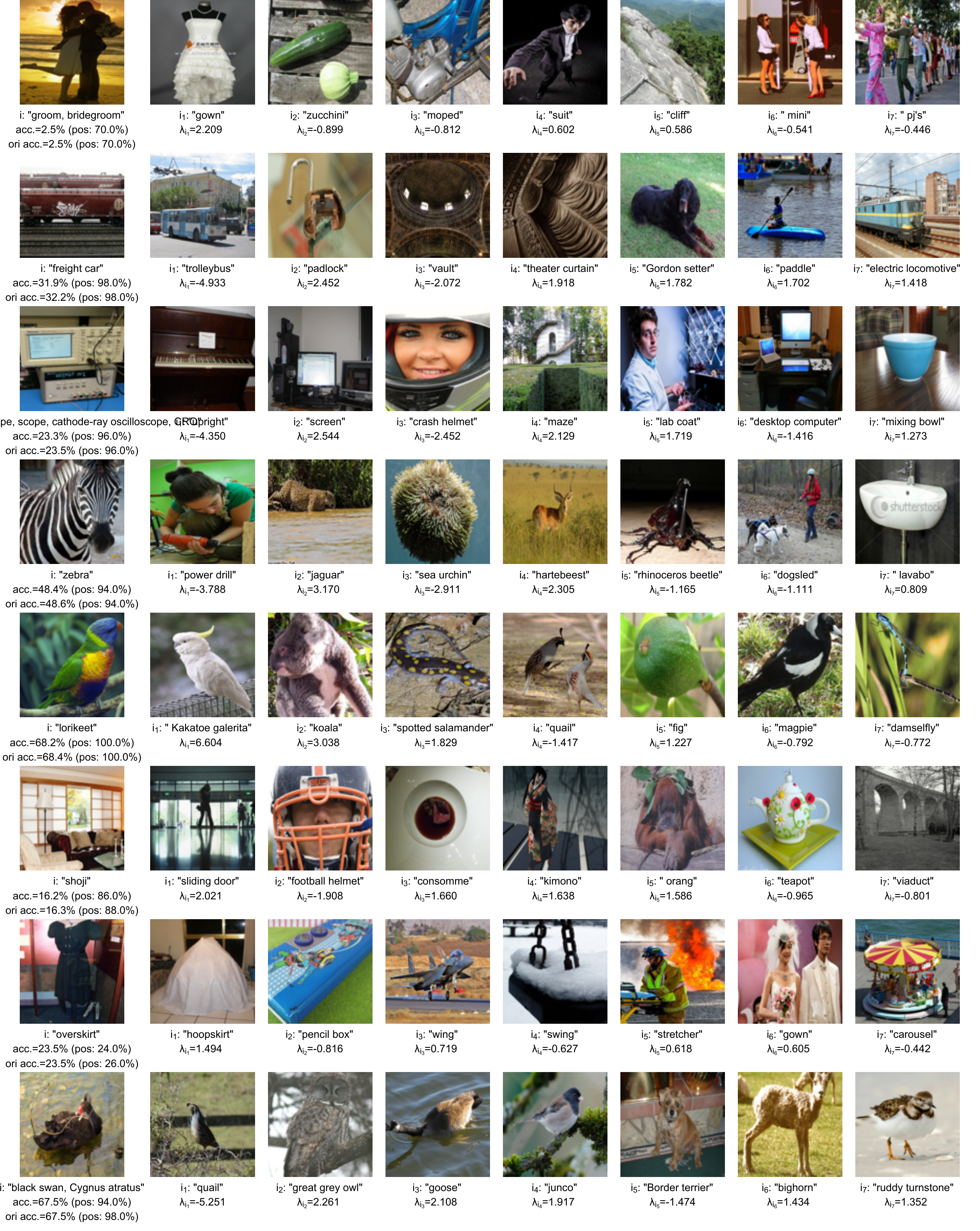}
\caption{Results from Swin-T$\to$ResNet-50, where `acc.' and `ori acc.' denote the classification accuracies on the ImageNet validation set, while `pos: xx\%' is the accuracy on positive samples only.}\label{fig:supp_deficit_swim_t_resnet50}
\end{figure}

\begin{figure}[t]
\centering
\includegraphics[width=0.98\linewidth]{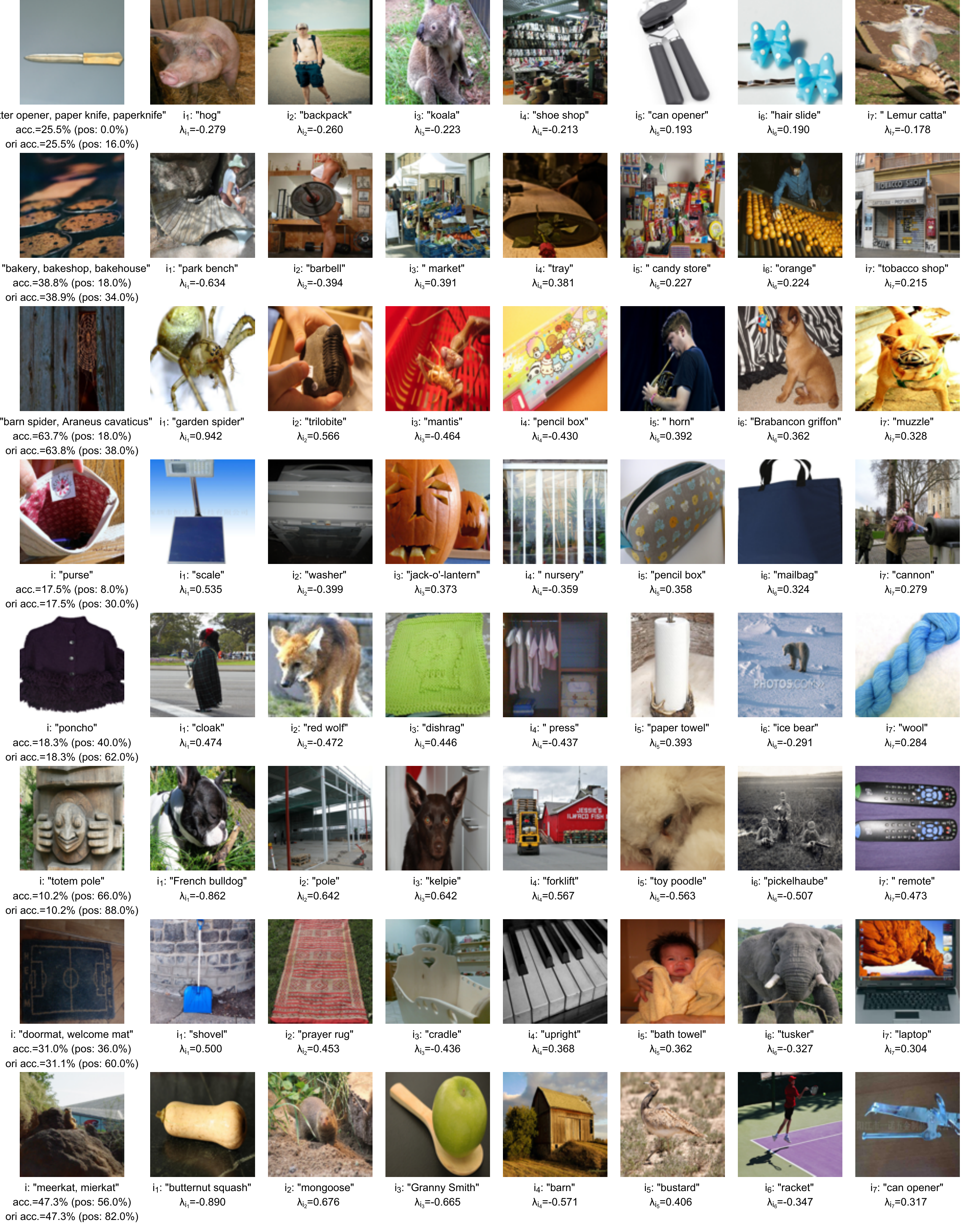}
\caption{Results from ResNet-50$\to$ResNet-18, where `acc.' and `ori acc.' denote the classification accuracies on the ImageNet validation set, while `pos: xx\%' is the accuracy on positive samples only.}\label{fig:supp_deficit_resnet50_resnet18}
\end{figure}

\begin{figure}[t]
\centering
\includegraphics[width=0.98\linewidth]{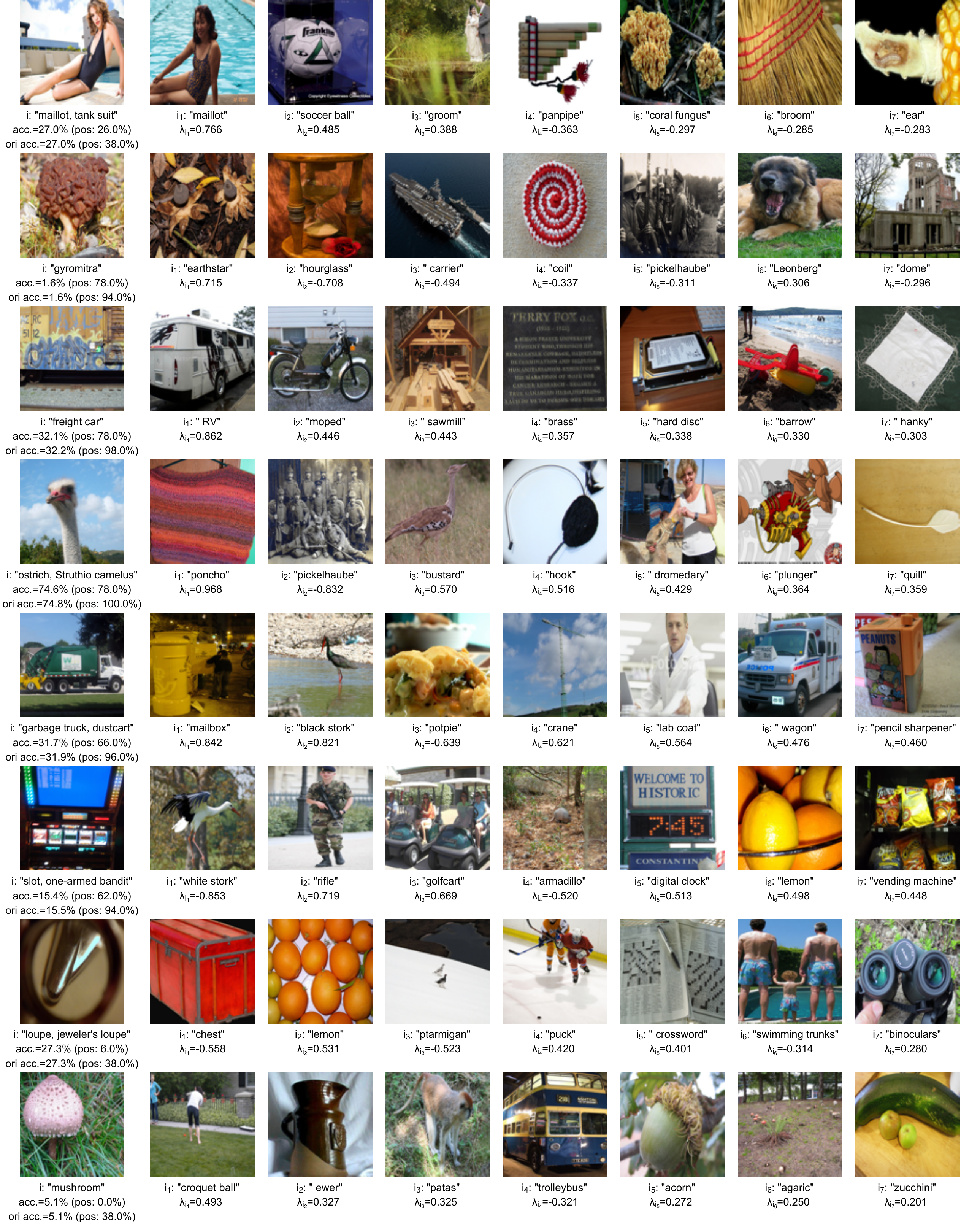}
\caption{Results from ResNet-18$\to$ResNet-50, where `acc.' and `ori acc.' denote the classification accuracies on the ImageNet validation set, while `pos: xx\%' is the accuracy on positive samples only.}\label{fig:supp_deficit_resnet18_resnet50}
\end{figure}

\begin{figure}[t]
\centering
\includegraphics[width=0.98\linewidth]{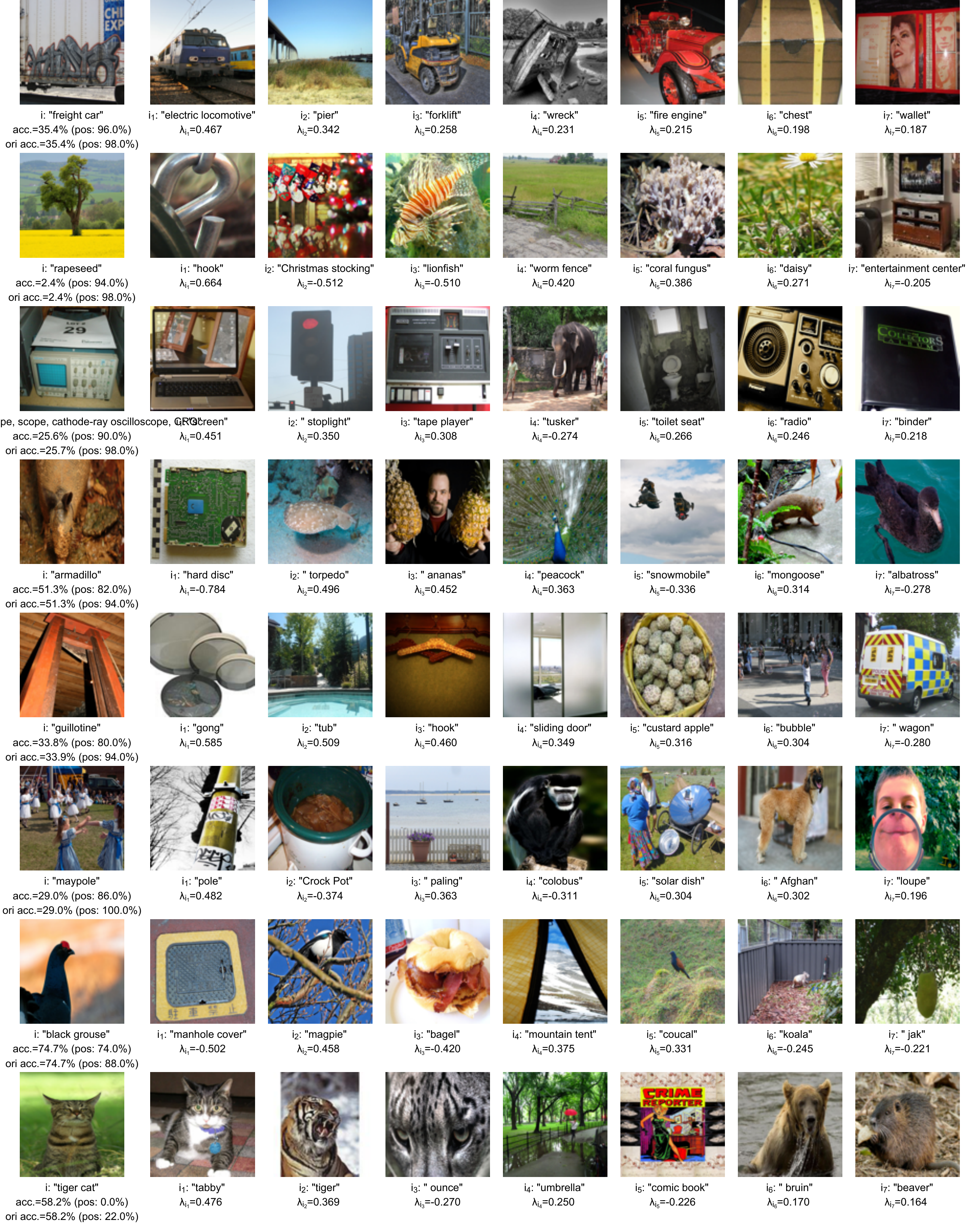}
\caption{Results from ResNet-50$\to$ViT-S, where `acc.' and `ori acc.' denote the classification accuracies on the ImageNet validation set, while `pos: xx\%' is the accuracy on positive samples only.}\label{fig:supp_deficit_resnet50_vit_s}
\end{figure}

\begin{figure}[t]
\centering
\includegraphics[width=0.98\linewidth]{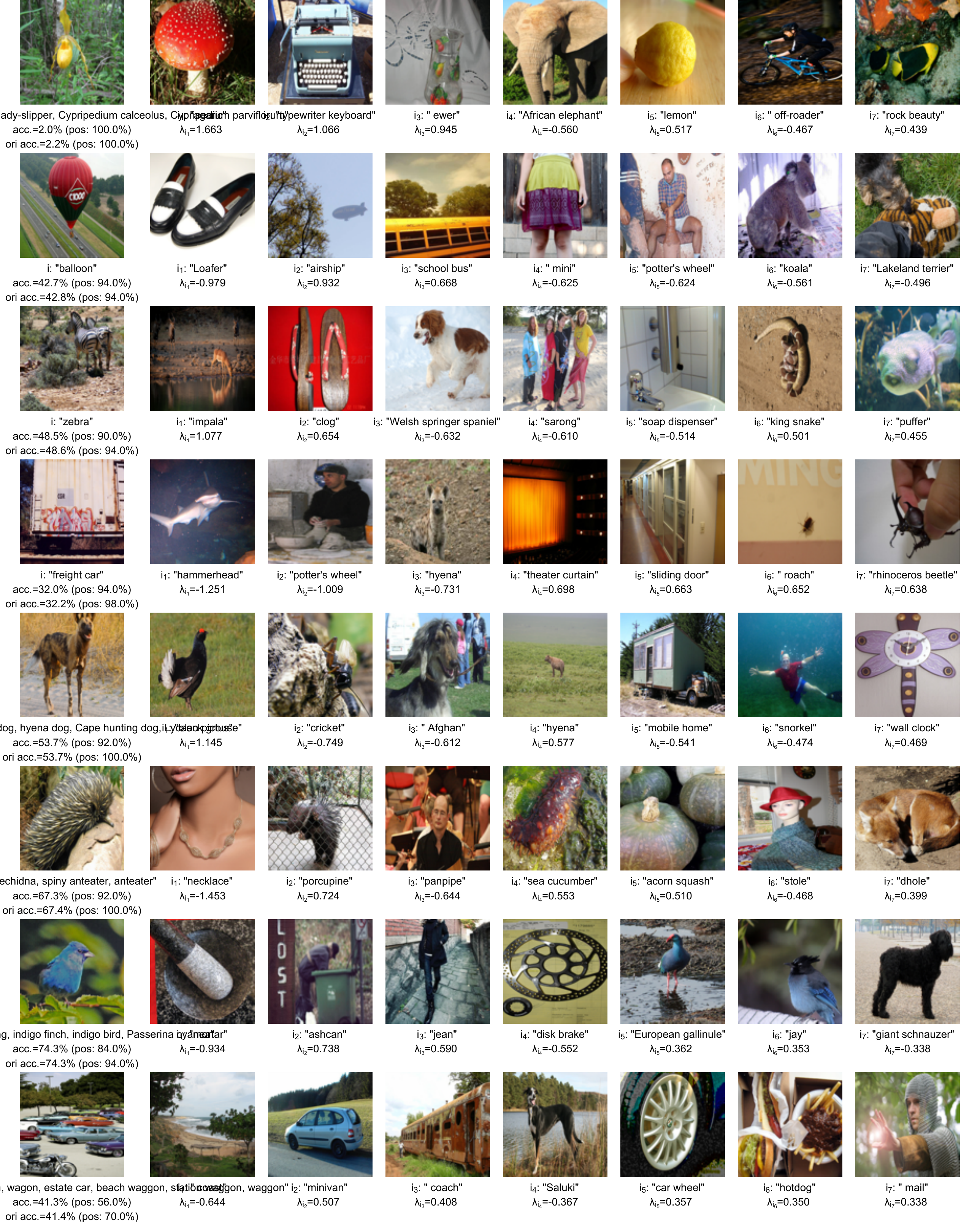}
\caption{Results from ViT-S$\to$ResNet-50, where `acc.' and `ori acc.' denote the classification accuracies on the ImageNet validation set, while `pos: xx\%' is the accuracy on positive samples only.}
\label{fig:supp_deficit_vis_s_resnet50}
\end{figure}

\begin{figure}[t]
\centering
\includegraphics[width=0.98\linewidth]{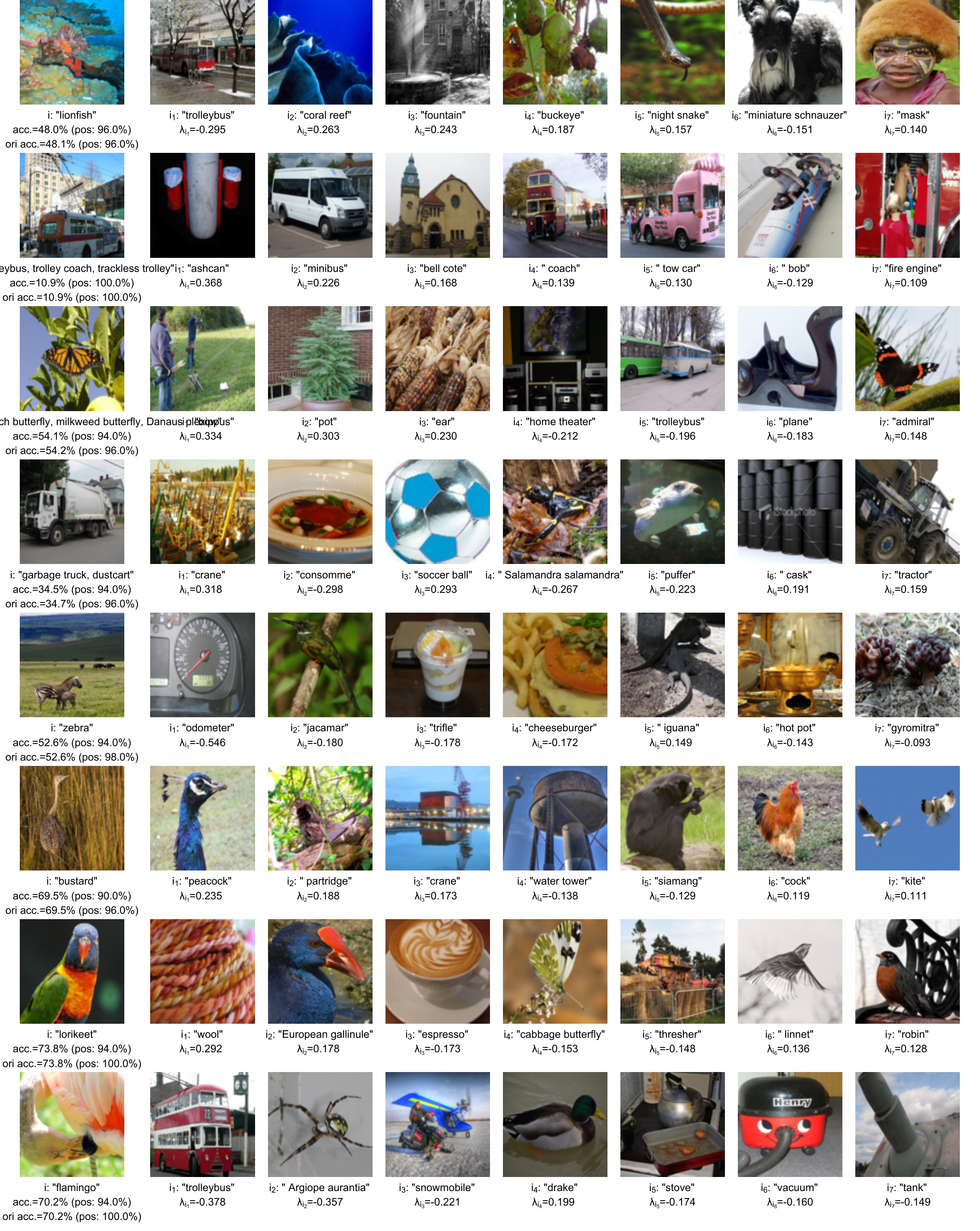}
\caption{Results from ResNet-18$\to$Swin-T, where `acc.' and `ori acc.' denote the classification accuracies on the ImageNet validation set, while `pos: xx\%' is the accuracy on positive samples only.}\label{fig:supp_deficit_resnet18_swim_t}
\end{figure}

\begin{figure}[t]
\centering
\includegraphics[width=0.98\linewidth]{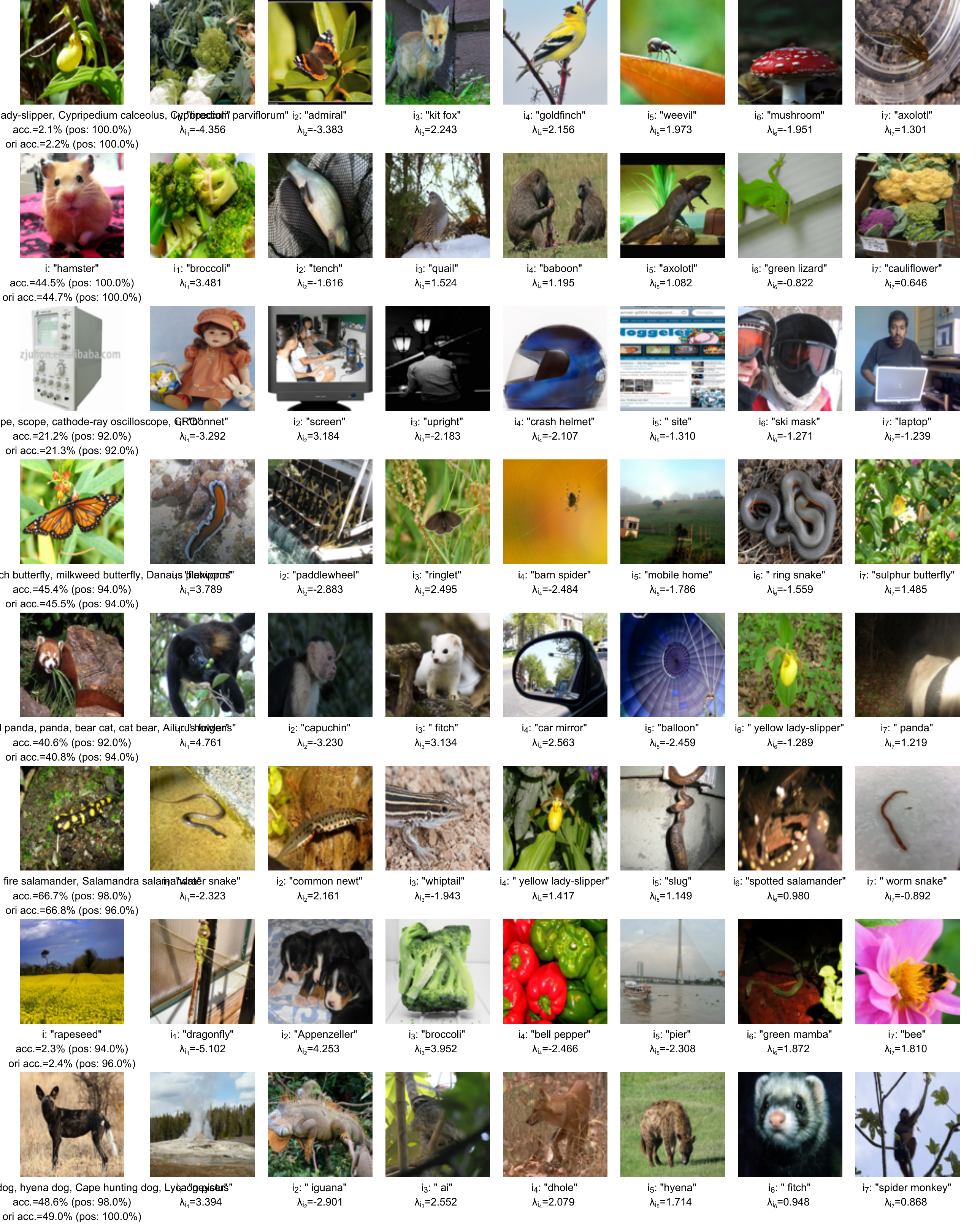}
\caption{Results from Swin-T$\to$ResNet-18, where `acc.' and `ori acc.' denote the classification accuracies on the ImageNet validation set, while `pos: xx\%' is the accuracy on positive samples only.}\label{fig:supp_deficit_swim_t_resnet18}
\end{figure}

\begin{figure}[t]
\centering
\includegraphics[width=0.98\linewidth]{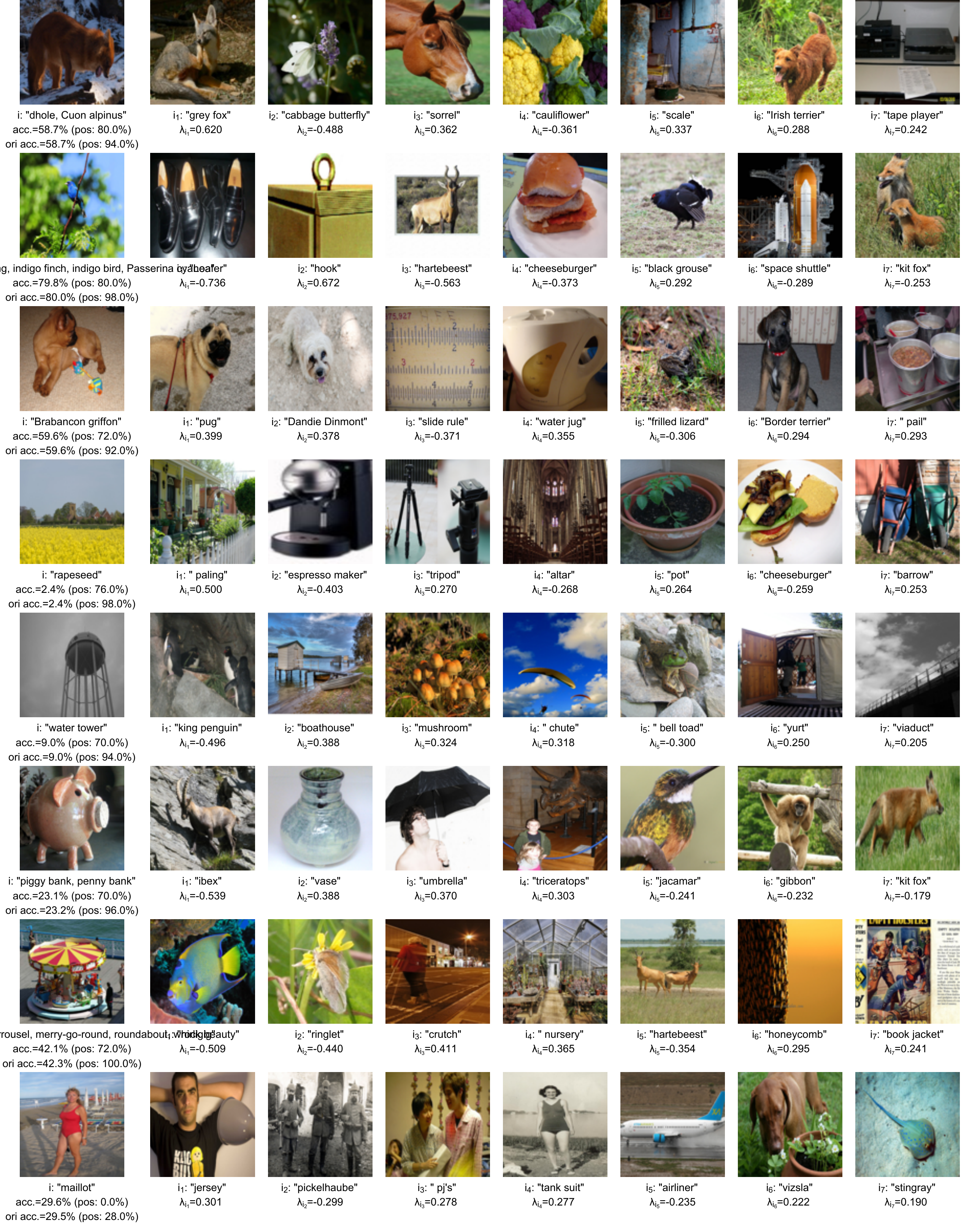}
\caption{Results from ResNet-50$\to$ViT-S, where `acc.' and `ori acc.' denote the classification accuracies on the ImageNet validation set, while `pos: xx\%' is the accuracy on positive samples only.}\label{fig:supp_deficit_resnet18_vit_t}
\end{figure}

\begin{figure}[t]
\centering
\includegraphics[width=0.98\linewidth]{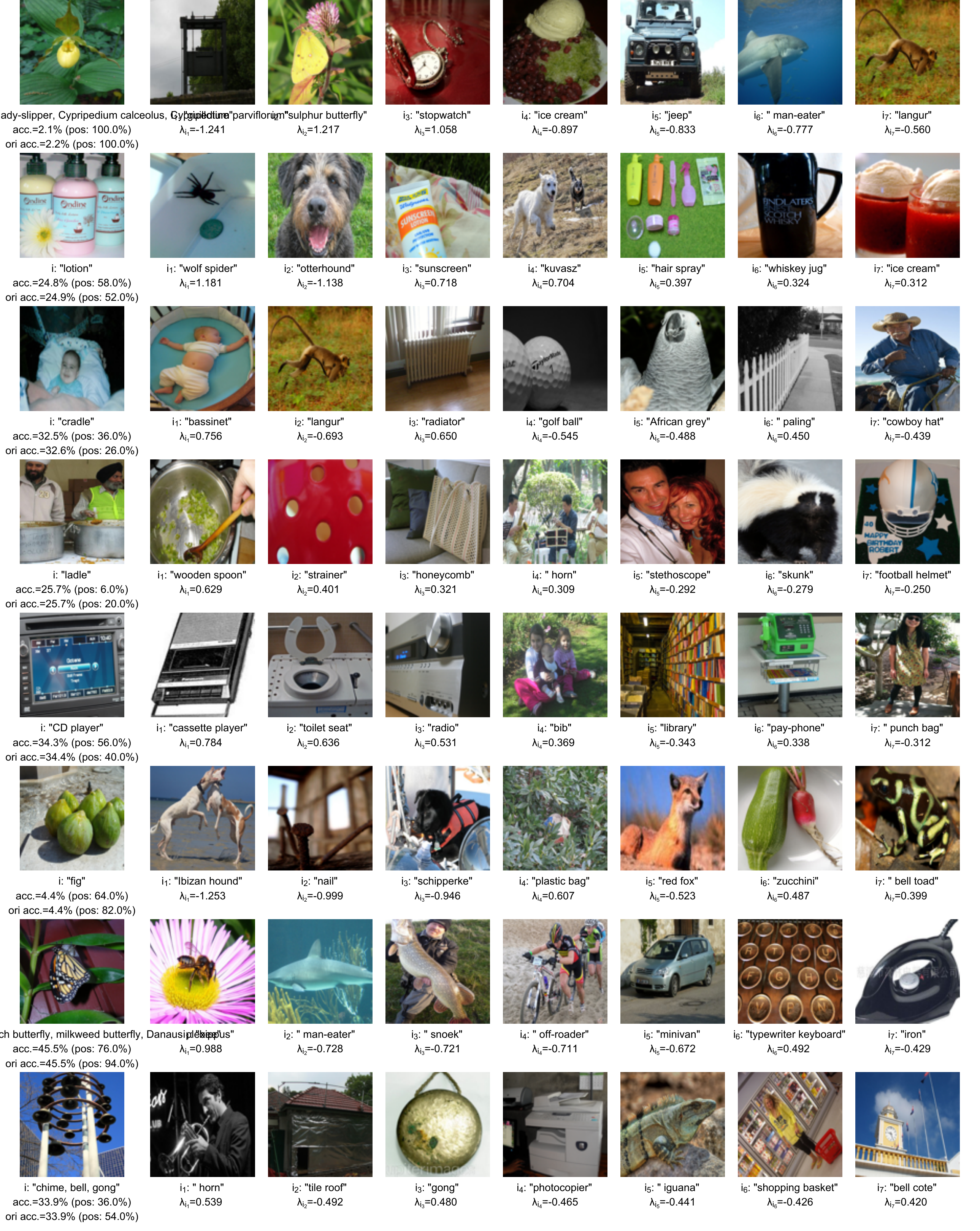}
\caption{Results from ViT-S$\to$ResNet-18, where `acc.' and `ori acc.' denote the classification accuracies on the ImageNet validation set, while `pos: xx\%' is the accuracy on positive samples only.}\label{fig:supp_deficit_vit_t_resnet50}
\end{figure}

\begin{figure}[t]
\centering
\includegraphics[width=0.98\linewidth]{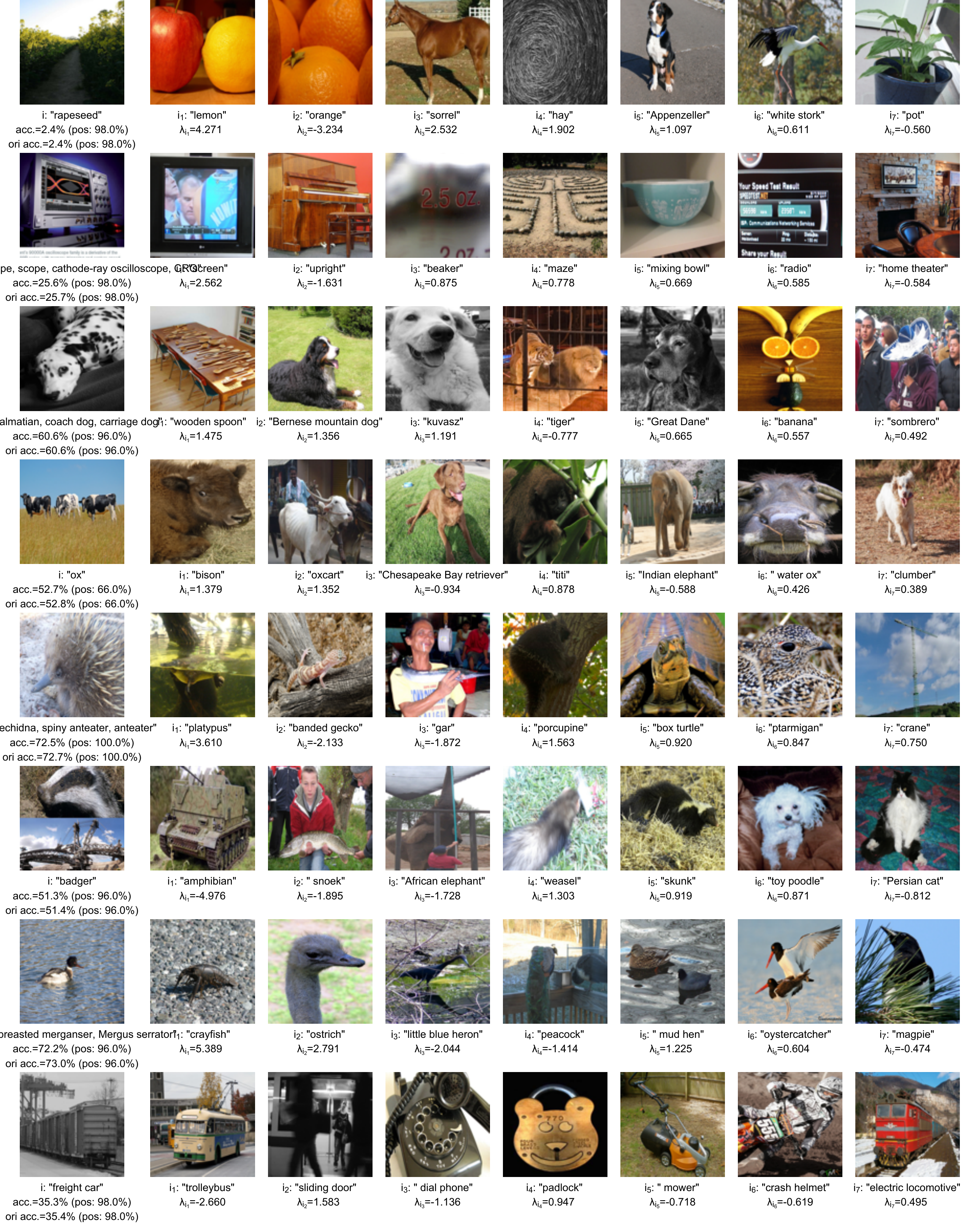}
\caption{Results from Swin-T$\to$ViT-S, where `acc.' and `ori acc.' denote the classification accuracies on the ImageNet validation set, while `pos: xx\%' is the accuracy on positive samples only.}\label{fig:supp_deficit_Swin-T_vit_t}
\end{figure}

\begin{figure}[t]
\centering
\includegraphics[width=0.98\linewidth]{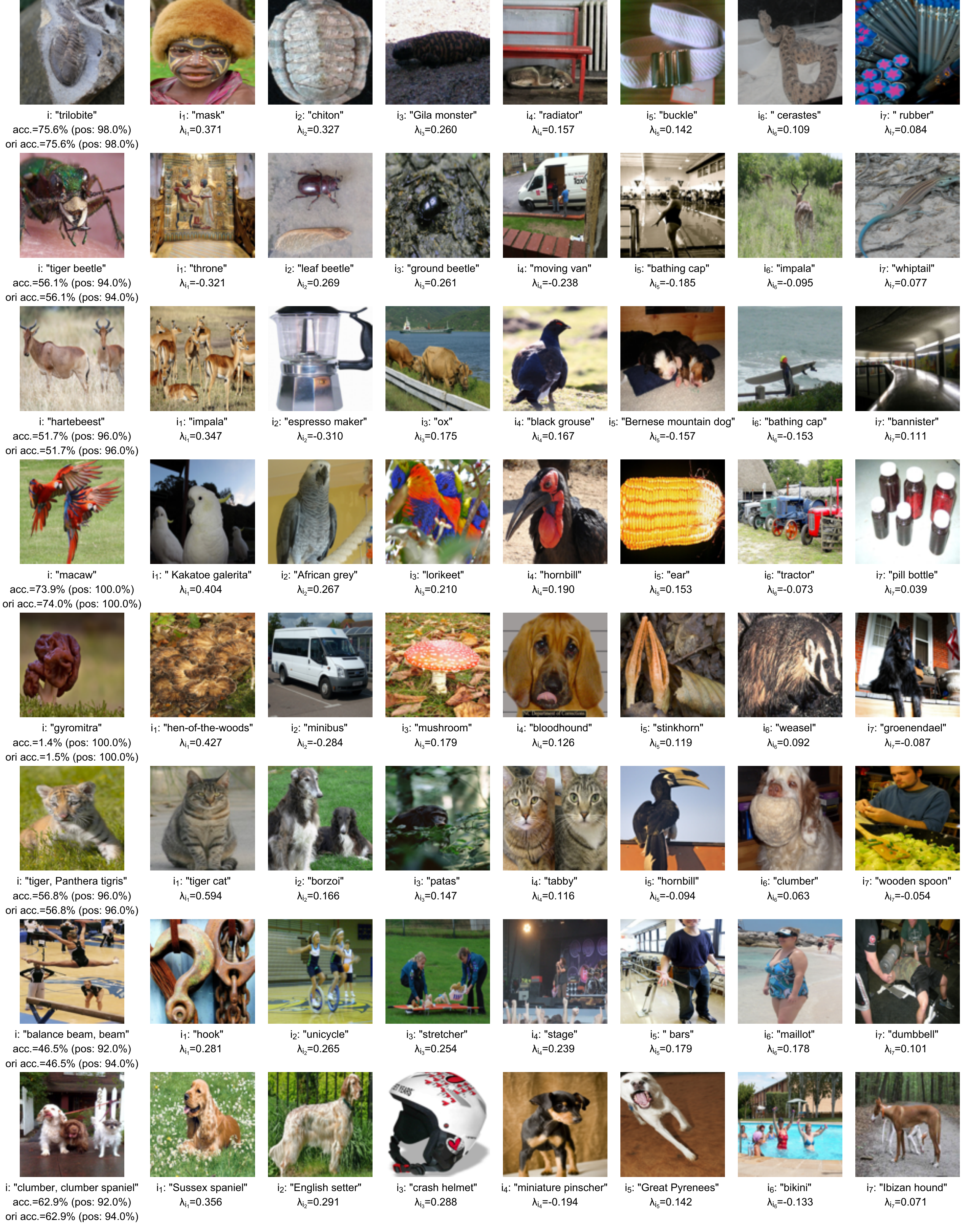}
\caption{Results from ViT-S$\to$Swin-T, where `acc.' and `ori acc.' denote the classification accuracies on the ImageNet validation set, while `pos: xx\%' is the accuracy on positive samples only.}\label{fig:supp_deficit_vit_s_Swin_T}
\end{figure}

We provide more examples of neural dependencies, which show that the logits predicted for some categories can be directly obtained by linearly combining the predictions of a few other categories. 
The results obtained by a single network (\ieno, ResNet-18, ResNet-50, ViT-S, and Swin-T) are reported in \cref{fig:supp_deficit_res18} - \cref{fig:supp_deficit_swin}, respectively. 
The results obtained between two independently-learned networks (\ieno, ResNet-18$\to$ResNet-50, ResNet-50$\to$ResNet-18, ResNet-18$\to$Swin-T, Swin-T$\to$ResNet-18, ResNet-18$\to$ViT-S, ViT-S$\to$ResNet-18, ResNet-50$\to$Swin-T, Swin-T$\to$ResNet-50, ResNet-50$\to$ViT-S, ViT-S$\to$ResNet-50, ViT-S$\to$Swin-T and ViT-S$\to$Swin-T) are reported in \cref{fig:supp_deficit_resnet50_swim_t} - \cref{fig:supp_deficit_vit_s_Swin_T}, respectively. 
All the results are obtained by solving the Lasso problem. In each figure, we report the classification accuracy for category `$i$': the accuracy by calculating logits is reported as `acc.'; the original model accuracy is reported as `ori. acc.'. Both metrics are measured in the whole ImageNet validation set. We further report the classification accuracy on positive samples only for both metrics as `pos' following `acc.' and `ori. acc.' correspondingly. The results show a neural independence phenomenon for broad categories in all those deep networks.



%%%%%%%%%%%%%%%%%%%%%%%%%cross model

\end{document}